\documentclass[runningheads]{llncs}
\usepackage{graphicx}
\usepackage{amsmath}
\usepackage{amssymb}
\usepackage{xcolor}
\begin{document}
	
	\title{Representation of preferences for multiple criteria decision aiding in a new seven-valued logic}
	
	\author{Salvatore Greco\inst{1} and Roman S{\l}owi\'{n}ski \inst{2}} 
	
	\institute{
		Deaprtment of Economics and Business, University of Catania,\\
		Corso Italia, 55, 95129 Catania, Italy \and
		Institute of Computing Science, Pozna\'{n} University of Technology,\\
		60-965 Pozna\'{n},
		and Systems Research Institute,\\
		Polish Academy of Sciences, 01-447 Warsaw, Poland\
	}
	
	\maketitle
	
	\begin{abstract}
		The seven-valued logic considered in this paper naturally arises within the rough set framework, allowing to distinguish vagueness due to imprecision from ambiguity due to coarseness. Recently, we discussed its utility for reasoning about data describing multi-attribute classification of objects. We also showed that this logic contains, as a particular case, the celebrated Belnap four-valued logic. Here, we present how the seven-valued logic, as well as the other logics that derive from it, can be used to represent preferences in the domain of Multiple Criteria Decision Aiding (MCDA). In particular, we propose new forms of outranking and value function preference models that aggregate multiple criteria taking into account imperfect preference information. We demonstrate that our approach effectively addresses common challenges in preference modeling for MCDA, such as uncertainty, imprecision, and ill-determination of performances and preferences. To this end, we present a specific procedure to construct a seven-valued preference relation and use it to define recommendations that consider robustness concerns by utilizing multiple outranking or value functions representing the decision maker's preferences. Moreover, we discuss the main properties of the proposed seven-valued preference structure and compare it with current approaches in MCDA, such as ordinal regression, robust ordinal regression, stochastic multiattribute acceptability analysis, stochastic ordinal regression, and so on. We illustrate and discuss the application of our approach using a didactic example. Finally, we propose directions for future research and potential applications of the proposed methodology. 
	\end{abstract}


	\section{Introduction}
	
	The seven-valued logic considered in this paper has been recently introduced by the authors in the context of rough-set-based reasoning about data \cite{GS_7v_2023} in order to distinguish vagueness due to imprecision from ambiguity due to coarseness. On the theoretical ground, we demonstrated that the Pawlak-Brouwer-Zadeh lattice is the proper algebraic structure for this seven-valued logic. We also showed that this logic contains, as a particular case, the celebrated Belnap four-valued logic \cite{Belnap77} applied to express preferences in Multiple Criteria Decision Aiding (MCDA) \cite{Tsoukias-Vincke}.

	It is worth noting that the seven-valued logic is interesting from a cognitive psychology perspective. According to the seminal article by Miller \cite{Miller1956}, entitled 'The Magical Number Seven, Plus or Minus Two: Some Limits on Our Capacity for Processing Information', it appears that individuals can effectively handle approximately seven stimuli simultaneously. This limit applies to both one-dimensional absolute judgment and short-term memory.
	
	To give an intuition of the seven-valued logic and the other logics deriving from it, let us consider the following example. Consider a hypothetical problem of evaluation of a finite set $\mathcal{A}$ of municipalities with respect to sustainable development. Suppose that three macrocriteria are considered for the evaluation of municipalities: economic $(Eco)$, social $(Soc)$, and environmental $(Env)$. Assume, moreover, that the overall evaluation of each municipality $a \in \mathcal{A}$, denoted by $U(a)$, is a weighted sum: 
	$$
	U(a) = w_{Eco}\times Eco(a) + w_{Soc}\times Soc(a) + w_{Env}\times Env(a),\ \  w_{Eco}+w_{Soc}+w_{Env}=1,
	$$
	and $w_{Eco}\ge 0,\ w_{Soc}\ge 0,\ w_{Env}\ge 0$. To consider the viewpoints of different stakeholders, three types of weight vectors, called \textit{perspectives}, are considered:
	\begin{itemize}
		\item \textit{Economic}, with $w_{Eco}>w_{Soc}=w_{Env}$,
		\item \textit{Social}, with $w_{Soc}>w_{Eco}=w_{Env}$,
		\item \textit{Environmental}, with $w_{Env}>w_{Eco}=w_{Soc}$.
	\end{itemize}
	It happens, however, that the stakeholders identified with a particular perspective, are not able to provide a precise values of the corresponding weight, that is $w_{Eco}$ for the economic perspective, $w_{Soc}$ for the social perspective, and $w_{Env}$ for the environmental perspective.  Instead, they agree to elicit some central values of the corresponding weights, satisfying the above constraints in each perspective. For example, if the central weight $w_{Eco}$ is set at 0.5, the other weights, $w_{Soc}$ and $w_{Env}$, are each set to 0.25. To make the evaluation more robust, the stakeholders agree to consider sets of weight vectors obtained by perturbation of the central weights within a given range of $r\%$, with a simultaneous adjustment of other weights, so that their sum equals always 1. Therefore, instead of a single overall evaluation in each perspective, each municipality $a \in \mathcal{A}$ gets a set of overall evaluations — including the central evaluation and a series of its `perturbations'. Let us denote by $\mathcal{U}^{Eco}(a),\ \mathcal{U}^{Soc}(a),$ and $\mathcal{U}^{Env}(a)$ the set of overall evaluations of $a \in \mathcal{A}$ in the economic, social and environmental perspectives, respectively.
	
	Evaluations related to one of the three perspectives will be denoted by $\mathcal{U}^p$, where $p$ can be $Eco$, $Soc$, or $Env$. Comparing municipality $a$ with municipality $b$  $(a,b\in\mathcal{A})$ in the considered perspective $p$, there are three possible situations: 
	\begin{itemize}
		\item $a$ is at least as good as $b$, because $a$ is at least as good as $b$ taking the central evaluation in perspective $p$ as well as all its `perturbations', that is, $U(a) \geq U(b)$ for all $U \in \mathcal{U}^{p}$,
		\item $a$ is not at least as good as $b$, because $a$ is worse than $b$ taking the central evaluation in perspective $p$ as well as  all its `perturbations', that is, $U(a) < U(b)$ for all $U \in \mathcal{U}^{p}$,
		\item it is unknown whether $a$ is at least as good as $b$, because $a$ is at least as good as  $b$ for some evaluations in perspective $p$ but worse for others, that is, $U(a) \geq U(b)$ for some $U \in \mathcal{U}^{p}$ and $U(a) < U(b)$ for some other $U \in \mathcal{U}^{p}$.
	\end{itemize}
	
	In result of the pairwise comparisons of municipality $a$ and municipality $b$ across the entire set of overall evaluations in all three perspectives, the proposition ``municipality $a$ is at least as good as municipality $b$'', denoted by $a \succsim b$, can assume one of the following seven possible states of truth: 
	\begin{itemize}
		\item $a$ is at least as good as $b$ in all three perspectives, that is, $a$ is at least as good as $b$ for all the evaluations in all three perspectives: then, proposition $a \succsim b$ is \textbf{true};  
		\item $a$ is at least as good as $b$ in one or two of the three perspectives, and it is unknown in the others, that is, $a$ is at least as good as $b$ for all the evaluations in one or two of the three perspectives, but there are evaluations for which $a$ is at least as good as $b$ and others for which this is not true in the remaining perspectives: then, proposition $a \succsim b$   is \textbf{sometimes true}; 
		\item it is unknown whether $a$ is at least as good as $b$ in all the three perspectives, that is, there are evaluations for which $a$ is at least as good as $b$ and others for which this is not true in all the three perspectives: then, proposition $a \succsim b$   is \textbf{unknown};   
		\item $a$ is at least as good as $b$ in one or two perspectives and this is false in the other perspectives, that is, $a$ is at least as good as $b$ for all the evaluations in one or two perspectives while this is false for all the evaluations in the other perspectives: then, proposition $a \succsim b$   is \textbf{contradictory};   
		\item  $a$ is at least as good as $b$ in one perspective, it is false in another perspective, and it is unknown in the remaining perspective, that is,  $a$ is at least as good as $b$ for all the evaluations in one perspective, it is false for all the evaluations in another perspective, and it is true for some evaluations and false for other evaluations in the remaining perspective: then, proposition $a \succsim b$   is  \textbf{fully contradictory};  
		\item $a$ is not at least as good as $b$ in one or two of the three perspectives and it is unknown in the other perspectives, that is, $a$ is not at least as good as $b$ for all the evaluations in one or two of the three perspectives, but there are evaluations for which $a$ is at least as good as $b$ and others for which this is not true in the remaining perspectives: then, proposition $a \succsim b$   is  \textbf{sometimes false};  
		\item $a$ is not at least as good as $b$ in all  the three perspectives, that is,  $a$ is not at least as good as $b$ for all the evaluations in all the three perspectives: then, proposition $a \succsim b$ is  \textbf{false}.  
	\end{itemize}
	
\begin{figure}[h]
		\centering
		\includegraphics[width=4cm]{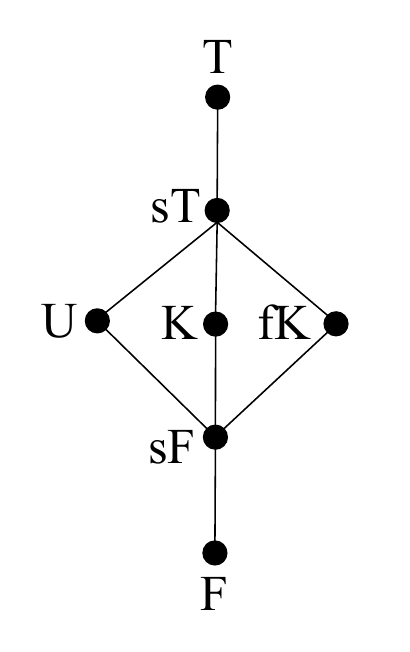}
		\caption{ \ Seven-valued logic truth value lattice }\label{Fig.1}
	\end{figure}

		The lattice presented in Figure \ref{Fig.1} illustrates the layered scheme of the truth values in the seven-valued logic, where higher layers represent greater certainty of truth.  

	The above seven cases are, of course, very detailed, so in particular decision situations it might be convenient to aggregate some of them for practical reasons. For example,  one could consider a bit less fine, but still quite detailed representation of preferences considering the following four-valued weak preference (for a discussion on the application of four-valued preference in multicriteria decision making see \cite{Tsoukias-Vincke}): 
	\begin{itemize}
		\item $a \succsim b$ is true if it is true or sometimes true in the above seven-valued weak preference relation; 
		\item $a \succsim b$ is unknown if it is unknown in the above seven-valued weak preference relation;
		\item $a \succsim b$ is contradictory if it is contradictory or fully contradictory in the above seven-valued preference relation; 
		\item $a \succsim b$ is false if it is false or sometimes false in the above seven-valued weak preference relation.
	\end{itemize}
	Another useful aggregation of the seven values of preference truth is the three-valued preference structure, derived from the above four-valued structure by combining the unknown, contradictory, and fully contradictory preference relations. Of course, other suitable preference structures can be created by different aggregations of the seven-valued preference relations.

	\vspace{6pt}
	In this paper, we take advantage of the seven-valued logic to handle robustness concerns in MCDA preference modeling. The paper is organized as follows. In the next Section, we sketch the presented methodology using block schemes representing its main steps. In Section 3, we explain the methodology with a didactic example. The last section groups conclusions. 
	
	\section{Main steps of the proposed methodology}

\begin{figure}[h]
		\centering
		\includegraphics[width=12cm]{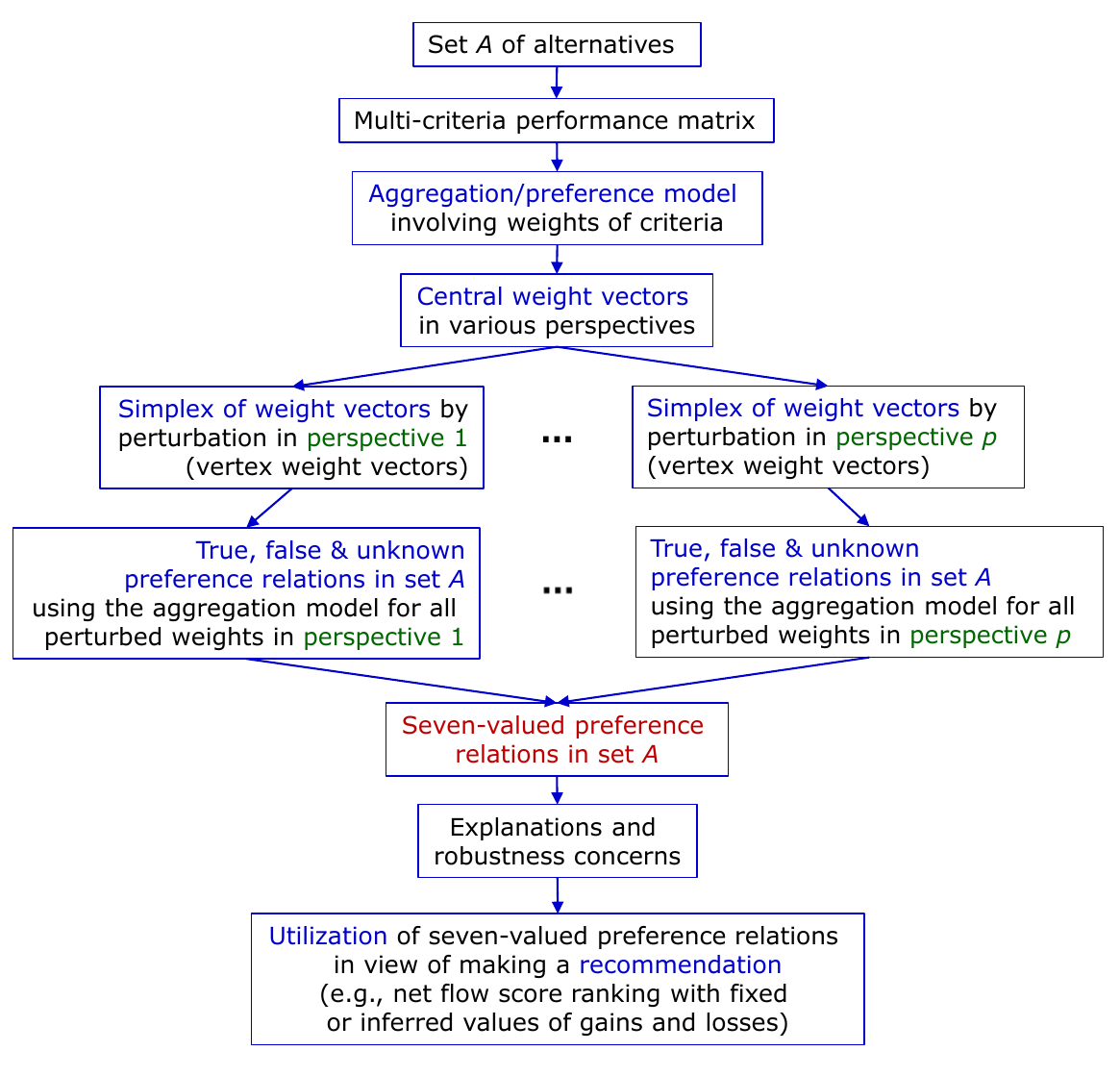}
		\caption{The methodology of construction of seven-valued preference relations and their utilization in view of making a ranking recommendation}\label{Fig.2}
	\end{figure}

\begin{figure}[h]
		\centering
		\includegraphics[width=12cm]{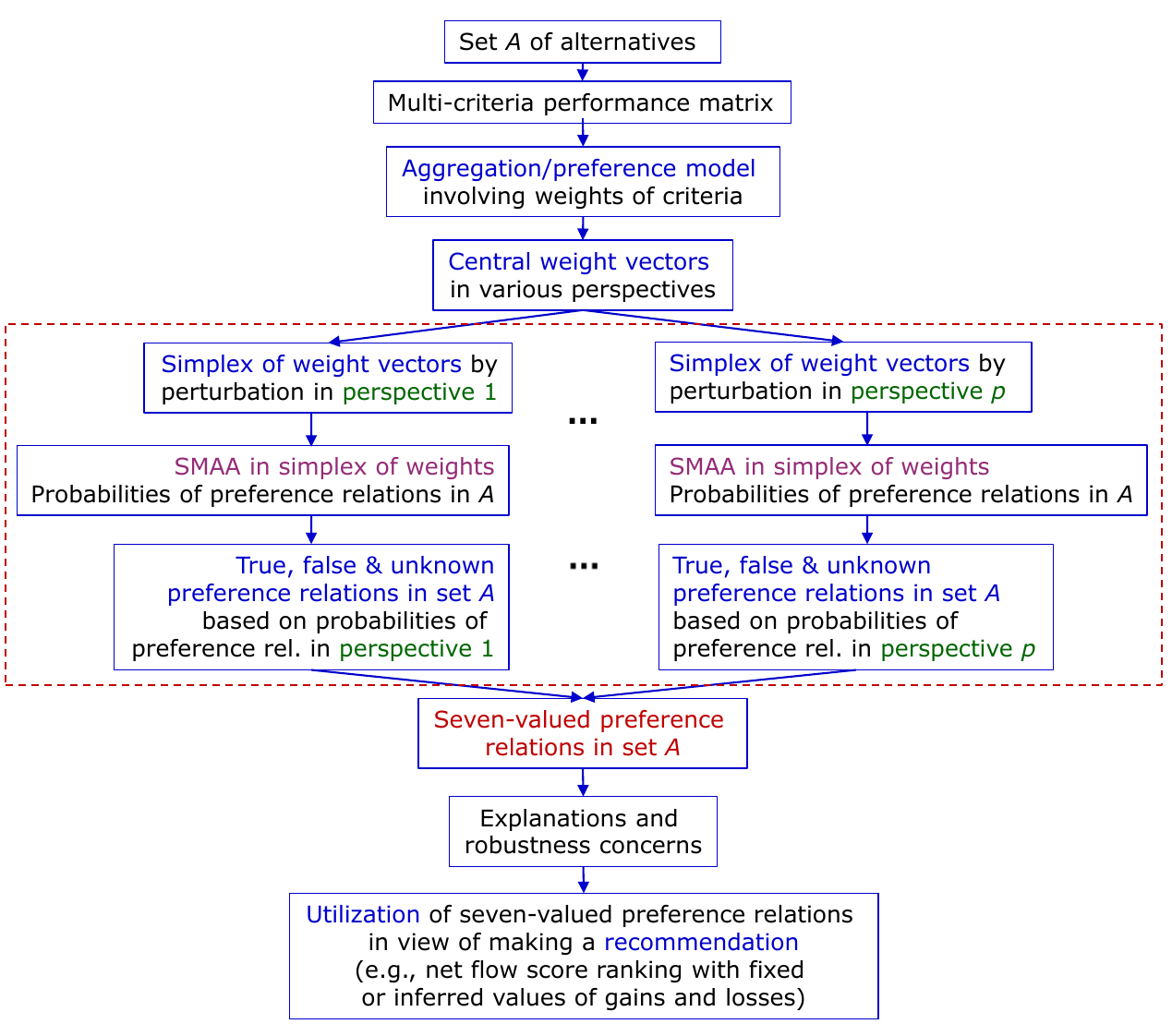}
		\caption{The first variant of the basic methodology - the changed part of the scheme is marked with a dashed line}\label{Fig.3}
	\end{figure}

\begin{figure}[h]
		\centering
		\includegraphics[width=12cm]{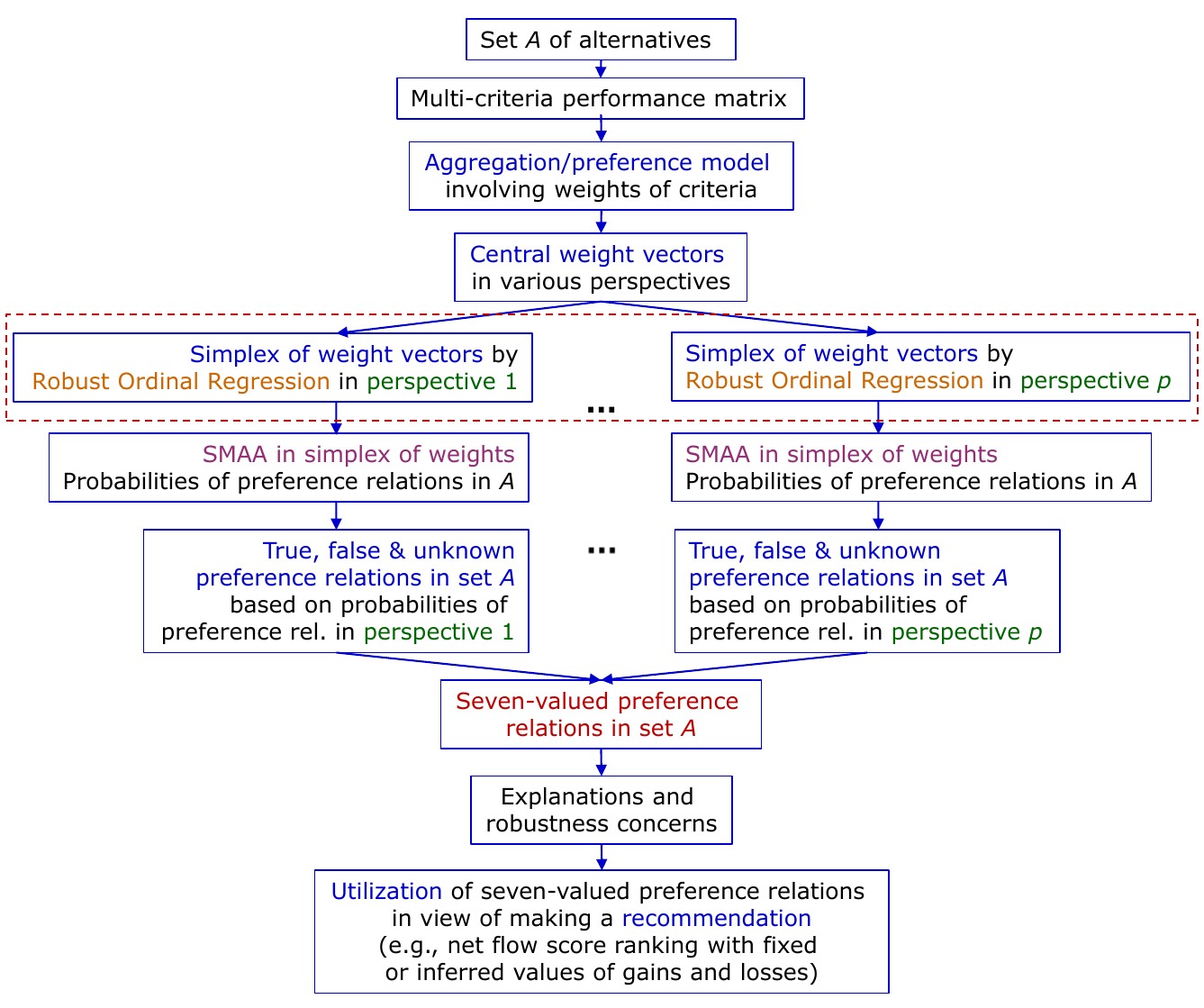}
		\caption{The second variant of the basic methodology - the changed part of the scheme is marked with a dashed line}\label{Fig.4}
	\end{figure}

	In this Section, we present the block schemes summarizing the proposed methodology (Figure \ref{Fig.2}) and its two variants (Figures \ref{Fig.3},\ref{Fig.4}). The variants concern the exploration of the space of feasible weights assigned to criteria. In the basic methodology sketched in Figure \ref{Fig.2}, the diversity of weight vectors in each perspective is obtained by a perturbation of central weights within the range of $r\%$. In the first variant of the methodology, presented in Figure \ref{Fig.3}, the space of feasible weights obtained by the perturbation is explored by SMAA (Stochastic Multiobjective Acceptability Analysis), providing probabilities of preference relations among alternatives, called pairwise winning indices. In the second variant of this methodology, presented in Figure \ref{Fig.4}, the space of feasible weights is obtained by ROR (Robust Ordinal Regression) on the base of holistic preference information provided by the Decision Maker (DM), and then this space is possibly explored by SMAA giving the probabilities of preference relations among alternatives (pairwise winning indices).  
	
	\section{Explaining the methodology with a didactic example}
\subsection{The didactic example}
	In this section, we are explaining step-by-step the methodology of multiple criteria decision aiding based on seven-valued representation of preferences using a didactic example.  Consider a dean who must compare five students, taking into account their grades in Mathematics $(Math)$, Physics $(Phys)$, Literature $(Lit)$, and Philosophy $(Phil)$. These grades, expressed on a scale from 0 to 100, are presented in Table \ref{Notes}.
	
	\begin{table}
		\caption{Grades of five students in Mathematics, Physics, Literature and Philosophy}\label{Notes}
		\centering\footnotesize
		\resizebox{0.6\textwidth}{!}{
			\begin{tabular}{l|cccc}
				Student  & Mathematics    & Physics & Literature & Philosophy \\
				\hline
				$S$1    & 80	& 90	& 50	& 70 \\
				$S$2          & 70	& 80	& 80	& 70 \\
				$S$3  & 100	& 60	& 50	& 70 \\
				$S$4      & 90	& 90	& 60	& 60 \\
				$S$5       & 80	& 80	& 70	& 70 \\
				\hline
			\end{tabular}
		}
	\end{table}
	
	Suppose a scenario where the dean begins comparing students using a value function $U: [0,100]^4 \rightarrow [0,100]$ assigning to each student $S$ the overall evaluation 
	$$
	U(Math(S),Phys(S),Lit(S),Phil(S))= 
	$$
	\vspace{-18pt}
	$$
	w_{Math}\times Math(S)+w_{Phis}\times Phys(S)+w_{Lit}\times Lit(S)+w_{Phil}\times Phil(S)
	$$
	with
	\begin{itemize}
		\item $Math(S),Phys(S),Lit(S)$ and $Phil(S)$ being the grades of student $S$ in Mathematics, Physics, Literature and Philosophy,
		respectively,
		\item $w_{Math}, w_{Phys}, w_{Lit}, w_{Phil}$, such that $w_{Math} \geq 0,\ w_{Phys} \geq 0,\ w_{Lit}\geq 0, \linebreak w_{Phil}\geq 0,\ w_{Math}+w_{Phys}+w_{Lit}+w_{Phil}=1$, being the weights of Mathematics, Physics, Literature and Philosophy, 	respectively. 
	\end{itemize}
	In this case, the weights $w_{Math}, w_{Phys}, w_{Lit}$ and $w_{Phil}$ represent the trade-offs between the grades of four subjects. These weights were determined using a procedure coherent with their intended meaning, such as SMART or SMARTER \cite{Edwards_Barron}.  
	For the sake of simplicity, we will denote the overall evaluation of student $S$ by value function $U$ as $U(S)$, instead of $U(Math(S),Phys(S),Lit(S),Phil(S))$. Using value function $U$ for comparing any two students $S, S'$, we conclude that $S$ is at least as good as $S'$ if $U(S) \geq U(S')$. Suppose, moreover, that the dean wants to evaluate the five students in thee different perspectives:
	\begin{itemize}
		\item an egalitarian perspective with respect to Sciences and Humanities, that is, Mathematics and Physics on one hand, and Literature and Philosophy on the other hand, so that equal weights are assigned to all the four subjects: then, $w^1_{Math}=w^1_{Phys}=w^1_{Lit}=w^1_{Phil}=0.25$;  
		\item an extreme perspective which gives a strong advantage to Sciences over Humanities, so that Mathematics and Physics are getting much larger weights than Literature and Philosophy: then, $w^2_{Math}=w^2_{Phys}=0.4$ and $w^2_{Lit}=w^2_{Phil}=0.1$;
		\item a moderate perspective, intermediate between the egalitarian and extreme perspectives, which gives a slight advantage to Sciences over Humanities, so that Mathematics and Physics are getting a bit larger weights than Literature and Philosophy: then, $w^3_{Math}=w^3_{Phys}=0.3$ and $w^3_{Lit}=w^3_{Phil}=0.2$.  
	\end{itemize}
	
	The overall evaluations of the five students by value functions representing the three perspectives are presented in Table \ref{Evaluations}.
	
	\begin{table}
		\caption{Overall evaluations of five students by value functions in the egalitarian, extreme and moderate perspectives}\label{Evaluations}
		\centering\footnotesize
		\resizebox{0.5\textwidth}{!}{
			\begin{tabular}{l|cccc}
				Student               & Egalitarian    & Extreme & Moderate &\\
				\hline
				$S$1    & 72.5	  & 80	& 75	 &\\
				$S$2    & 75	  & 75	& 75	 &\\
				$S$3    & 70	  & 76	& 72	 &\\
				$S$4    & 75	  & 84	& 78	 &\\
				$S$5    & 75	  & 78	& 76	 &\\
				\hline
			\end{tabular}
		}
	\end{table}
	
	Looking at Table \ref{Evaluations}, one can note that 
	\begin{itemize}
		\item $S1$ has a better evaluation than $S3$ in all three perspectives,
		\item $S4$ has a not worse evaluation than all other students in all three perspectives,
		\item $S5$ has a not worse evaluation than $S2$ and $S3$ in all three perspectives,
		\item for all other pairs of students there is no definite preference in all three perspectives, because for each pair $S, S'$, student $S$ is better than $S'$ in some perspective, and student $S'$ is better than $S$ in some other perspective.
	\end{itemize}
	
	\subsection{Construction of the seven-valued preference relations with value function aggregation}
	The dean aims to address robustness concerns by studying how overall evaluations might change if the original weights, which we will call \textit{central} weights, for all three perspectives were perturbed within the range $r$, such as 15\%. Consequently, for each of the three perspectives, the perturbed weight vectors 
	$$\mathbf{\widetilde{w}}^p=[\widetilde{w}^p_{Math},\widetilde{w}^p_{Phys},\widetilde{w}^p_{Lit},\widetilde{w}^p_{Phil}],$$
	$p=1, 2, 3$, satisfying the following set of constraints, are considered:

	\begin{equation}\label{perturbation}
		\begin{array}{l}
			\left.
			\begin{array}{l}
				\widetilde{w}^p_{Math} \geq 0, \widetilde{w}^p_{Phys} \geq 0, \widetilde{w}^p_{Lit}\geq 0, \widetilde{w}^p_{Phil}\geq 0,\\[1mm]
				\widetilde{w}^p_{Math}+\widetilde{w}^p_{Phys}+\widetilde{w}^p_{Lit}+\widetilde{w}^p_{Phil}=1,\\[1mm]
				w^p_{Math}(1-r) \leq \widetilde{w}^p_{Math} \leq w^p_{Math}(1+r),\\[1mm]
				w^p_{Phys}(1-r) \leq \widetilde{w}^p_{Phys} \leq w^p_{Phys}(1+r),\\[1mm]
				w^p_{Lit}(1-r) \leq \widetilde{w}^p_{Lit} \leq w^p_{Lit}(1+r),\\[1mm]
				w^p_{Phil}(1-r) \leq \widetilde{w}^p_{Phil} \leq w^p_{Phil}(1+r).\\[1mm]
			\end{array}
			\right\}E^p_{ (weight \; perturbation) }
		\end{array}\nonumber
	\end{equation} 
	
	The overall evaluation of student $S$ by the value function with weight vector $\mathbf{\widetilde{w}}^p$ is denoted by $U(S,\mathbf{\widetilde{w}}^p),\ p=1,2,3$,  that is:
	\[
	U(S,\mathbf{\widetilde{w}}^p)=\widetilde{w}^p_{Math}\times Math(S)+\widetilde{w}^p_{Phys}\times Phys(S)+\widetilde{w}^p_{Lit}\times Lit(S)+\widetilde{w}^p_{Phil}\times Phil(S).
	\]
	
	Taking into account the perturbed weights in one perspective $p \in \{1,2,3\}$, we conclude that the proposition ``student $S$ is at least as good as student $S'$'' is:
	\begin{itemize}
		\item  true, and denoted by $S \succsim^{p,T} S'$, if $U(S,\mathbf{\widetilde{w}}^p) \geq U(S',\mathbf{\widetilde{w}}^p)$ for all $\mathbf{\widetilde{w}}^p$ satisfying the constraints $E^p_{ (weight \; perturbation) }$;
		\item false, and denoted by $S \succsim^{p,F} S'$, if $U(S,\mathbf{\widetilde{w}}^p) < U(S',\mathbf{\widetilde{w}}^p)$ for all $\mathbf{\widetilde{w}}^p$ satisfying the constraints $E^p_{ (weight \; perturbation) }$;
		\item unknown, and denoted by $S \succsim^{p,U} S'$, if $U(S,\mathbf{\widetilde{w}}^p) \geq U(S',\mathbf{\widetilde{w}}^p)$ for some $\mathbf{\widetilde{w}}^p$ satisfying the constraints $E^p_{ (weight \; perturbation) }$ and $U(S,\mathbf{\widetilde{w}}^p) < U(S',\mathbf{\widetilde{w}}^p)$ for some other $\mathbf{\widetilde{w}}^p$ satisfying the same constraints.
	\end{itemize}
	
	Taking into account the perturbed weights in all three perspectives $p=1,2,3$, we conclude that the proposition ``student $S$ is at least as good as student $S'$'' is: 
	\begin{itemize}
		\item \textbf{true}, and denoted by $S \succsim^{T} S'$, if $S \succsim^{p,T} S'$ for $p=1,2,3$;
		\item \textbf{sometimes true}, and denoted by $S \succsim^{sT} S'$, if $S \succsim^{p,T} S'$ in one or two perspectives $p\in \{1,2,3\}$ and $S \succsim^{p,U} S'$ in another perspective $p$;
		\item \textbf{unknown}, and denoted by $S \succsim^{U} S'$, if $S \succsim^{p,U} S'$ for $p=1,2,3$;
		\item \textbf{contradictory}, and denoted by $S \succsim^{K} S'$, if $S \succsim^{p,T} S'$ in one or two perspectives $p\in \{1,2,3\}$ and $S \succsim^{p,F} S'$ in another perspective $p$;
		\item \textbf{fully contradictory}, and denoted by $S \succsim^{fK} S'$, if $S \succsim^{p,T} S'$ in one perspective $p\in \{1,2,3\}$, $S \succsim^{p,F} S'$ in another perspective $p$, and $S \succsim^{p,U} S'$ in the remaining perspective $p$;
		\item \textbf{sometimes false}, and denoted by $S \succsim^{sF} S'$, if $S \succsim^{p,F} S'$ in one or two perspectives $p\in \{1,2,3\}$ and $S \succsim^{p,U} S'$ in another perspective $p$;
		\item \textbf{false}, and denoted by $S \succsim^{F} S'$, if $S \succsim^{p,F} S'$ for $p=1,2,3$.
	\end{itemize}
	
	To simplify notation, let us denote the set of all weight vectors $\mathbf{\widetilde{w}}^p$ satisfying the constraints $E^p_{ (weight \; perturbation)}$  by $E^p_{ (wp)}$. Clearly, $E^p_{ (wp)}$ is a convex polyhedron in $\mathbb{R}^4$ and the points of $E^p_{ (wp)}$ are all and only the convex combinations of its vertices. More precisely, denoting the set of vertices of $E^p_{ (wp)}$ by $V(E^p_{ (wp)})$, for all $\mathbf{\widetilde{w}}^p \in E^p_{ (wp)}$, we have:
	\[
	\mathbf{\widetilde{w}}^p=\sum_{{\mathbf{\widehat{w}}^p}\in V(E^p_{ (wp)})} \alpha_{\mathbf{\widehat{w}}^p}\times \mathbf{\widehat{w}}^p
	\]
with $\alpha_{{\mathbf{\widehat{w}}}^p} \geq 0$ for all vertices $\mathbf{\widehat{w}}^p \in V(E^p_{(wp)})$ and $\sum_{{\mathbf{\widehat{w}}^p}\in V(E^p_{ (wp)})} \alpha_{\mathbf{\widehat{w}}^p}=1$. 

To compute the preference relations $\succsim^{p,H}, H \in \{T,F,U\}$, in each particular perspective $p\in\{1,2,3\}$, and, on this basis, the overall seven-valued preference relations $\succsim^{K},\ K\in\{T,sT,U,K,fK,sF,F\}$, the following two propositions are useful. \\

\textbf{Proposition 1.} \textit{For all pairs of students, $S$ and $S'$, and constraints $E^p_{ (wp)}$ on perturbed weight vectors in one perspective $p \in \{1,2,3\}$, it holds that:
\begin{itemize}
	\item $S \succsim^{p,T} S'$ if and only if  $m^p(S,S') \geqslant 0$,
	\item $S \succsim^{p,F} S'$ if and only if  $M^p(S,S') < 0$,
	\item $S \succsim^{p,U} S'$, if and only if $m^p(S,S') < 0 \leqslant M^p(S,S')$,
	\end{itemize}
	with
	\begin{itemize}
		\item $m^p(S,S')=min[ U(S)-U(S')]$ subject to $E^p_{ (wp)}$,
		\item $M^p(S,S')=max [U(S)-U(S')]$ subject to $E^p_{ (wp)}$.
	\end{itemize}}
The proof can be found in Appendix A.

\vspace{6pt}

\textbf{Proposition 2.} \textit{For all pairs of students, $S$ and $S'$, and constraints $E^p_{ (wp)}$ on perturbed weight vectors in one perspective $p \in \{1,2,3\}$,  it holds that:
	\begin{itemize}
		\item $S \succsim^{p,T} S'$ if and only if  $U(S,\mathbf{\widetilde{w}}^p) \geq U(S',\mathbf{\widetilde{w}}^p)$ for all $\mathbf{\widetilde{w}}^p \in V(E^p_{ (wp)})$,
		\item $S \succsim^{p,F} S'$ if and only if  $U(S,\mathbf{\widetilde{w}}^p) < U(S',\mathbf{\widetilde{w}}^p)$ for all $\mathbf{\widetilde{w}}^p \in V(E^p_{ (wp)})$,
		\item $S \succsim^{p,U} S'$ if and only if $U(S,\mathbf{\widetilde{w}}^p) \geq U(S',\mathbf{\widetilde{w}}^p)$ for some $\mathbf{\widetilde{w}}^p \in V(E^p_{ (wp)})$ and $U(S,\mathbf{\widetilde{w}}^p) < U(S',\mathbf{\widetilde{w}}^p)$ for some other $\mathbf{\widetilde{w}}^p \in V(E^p_{ (wp)})$.
	\end{itemize}
}
The proof can be found in Appendix B.
\vspace{6pt}

%

In Tables \ref{Ma_mi_egalitarian}, \ref{Ma_mi_extreme} and \ref{Ma_mi_moderate}, we present the results of the application of Proposition 1, i.e., the values of $m^p(S,S')$ and $M^p(S,S')$, and the resulting preference relations $\succsim^{p,H},\ H \in \{T,F,U\}$, in each particular perspective $p\in \{1,2,3\}$, respectively.

\begin{table}
\caption{Values of $m^1(S,S')$ and $M^1(S,S')$ (in parenthesis), and resulting preference relations between students in the egalitarian perspective and value function aggregation: $\succsim^{1,T}$, $\succsim^{1,F}$, and $\succsim^{1,U}$}\label{Ma_mi_egalitarian}
\centering\footnotesize
\resizebox{1\textwidth}{!}{
\begin{tabular}{l|c|c|c|c|c}
Student               & $S$1    & $S$2 & $S$3 & $S$4 & $S$5\\
\hline
$S$1    & (0,0)$\rightarrow \succsim^{1,T}$	  & (-4.375,-0.625)$\rightarrow \succsim^{1,F}$	& (0.625,  4.375)$\rightarrow \succsim^{1,T}$	& (-3.625,-1.375)$\rightarrow \succsim^{1,F}$ & (-3.625,-1.375)$\rightarrow \succsim^{1,F}$ \\
$S$2    & (0.625,4.375)$\rightarrow\succsim^{1,T}$	  & (0,0)$\rightarrow\succsim^{1,T}$	& (2,8)$\rightarrow\succsim^{1,T}$ & (-2.25,2.25)$\rightarrow\succsim^{1,U}$	& (-0.75,0.75)$\rightarrow\succsim^{1,U}$ \\
$S$3    & (-4.375,-0.625)$\rightarrow\succsim^{1,F}$	  & (-8,-2)$\rightarrow\succsim^{1,F}$	& (0,0)$\rightarrow\succsim^{1,T}$ & (-7.25,-2.75)$\rightarrow\succsim^{1,F}$ & (-7.25,-2.75)$\rightarrow\succsim^{1,F}$	 \\
$S$4    & (1.375,3.625)$\rightarrow\succsim^{1,T}$	  & (-2.25,2.25)$\rightarrow\succsim^{1,U}$	& (2.75,7.25)$\rightarrow\succsim^{1,T}$	& (0,0)$\rightarrow\succsim^{1,T}$ & (-1.5,1.5 )$\rightarrow\succsim^{1,U}$\\
$S$5    & (1.375,3.625)$\rightarrow\succsim^{1,T}$	  & (-0.75,0.75)$\rightarrow\succsim^{1,U}$	& (2.75,7.25)$\rightarrow\succsim^{1,T}$	& (-1.5,1.5)$\rightarrow\succsim^{1,U}$ & (0,0)$\rightarrow\succsim^{1,T}$\\
\hline
\end{tabular}
}
\end{table}


\begin{table}
\caption{Values of $m^2(S,S')$ and $M^2(S,S')$ (in parenthesis), and resulting preference relations between students in the extreme perspective and value function aggregation: $\succsim^{2,T}$, $\succsim^{2,F}$, and $\succsim^{2,U}$}\label{Ma_mi_extreme}
\centering\footnotesize
\resizebox{1\textwidth}{!}{
\begin{tabular}{l|c|c|c|c|c}
Student               & $S$1    & $S$2 & $S$3 & $S$4 & $S$5\\
\hline
$S$1    & (0,0)$\rightarrow\succsim^{2,T}$	  & (4.25,5.75)$\rightarrow\succsim^{2,T}$	& (1,7)$\rightarrow\succsim^{2,T}$	& (-4.9,-3.1)$\rightarrow\succsim^{2,F}$ & (1.1,2.9)$\rightarrow\succsim^{2,T}$ \\
$S$2    & (-5.75,-4.25)$\rightarrow\succsim^{2,F}$	  & (0,0)$\rightarrow\succsim^{2,T}$	& (-4.45,2.45)$\rightarrow\succsim^{2,U}$ & (-10.35,-7.65)$\rightarrow\succsim^{2,F}$	& (-3.75,-2.25)$\rightarrow\succsim^{2,F}$ \\
$S$3    & (-7,-1)$\rightarrow\succsim^{2,F}$	  & (-2.45,4.45)$\rightarrow\succsim^{2,U}$	& (0,0)$\rightarrow\succsim^{2,T}$ & (-10.7,-5.3)$\rightarrow\succsim^{2,F}$ & (-4.7,0.7)$\rightarrow\succsim^{2,U}$	 \\
$S$4    & (3.1,4.9)$\rightarrow\succsim^{2,T}$	  & (7.65,10.35)$\rightarrow\succsim^{2,T}$	& (5.3,10.7)$\rightarrow\succsim^{2,T}$	& (0,0)$\rightarrow\succsim^{2,T}$ & (5.4,6.6)$\rightarrow\succsim^{2,T}$\\
$S$5    & (-2.9,-1.1)$\rightarrow\succsim^{2,F}$	  & (2.25,3.75)$\rightarrow\succsim^{2,T}$	& (-0.7,4.7)$\rightarrow\succsim^{2,U}$	& (-6.6,-5.4)$\rightarrow\succsim^{2,F}$ & (0,0)$\rightarrow\succsim^{2,T}$\\
\hline
\end{tabular}
}
\end{table}


\begin{table}
\caption{Values of $m^3(S,S')$ and $M^3(S,S')$ (in parenthesis), and resulting preference relations between students in the moderate perspective and value function aggregation: $\succsim^{3,T}, \succsim^{3,F}$ and $\succsim^{3,U}$}\label{Ma_mi_moderate}
\centering\footnotesize
\resizebox{1\textwidth}{!}{
\begin{tabular}{l|c|c|c|c|c}
Student               & $S$1    & $S$2 & $S$3 & $S$4 & $S$5\\
\hline
$S$1    & (0,0)$\rightarrow\succsim^{3,T}$	  & (-1.5,1.5)$\rightarrow\succsim^{3,U}$	& (0.75,5.25)$\rightarrow\succsim^{3,T}$	&(-4.05,-1.95)$\rightarrow\succsim^{3,F}$ & (-2.05,0.05)$\rightarrow\succsim^{3,U}$ \\
$S$2    & (-1.5,1.5)$\rightarrow\succsim^{3,U}$	  & (0,0)$\rightarrow\succsim^{3,T}$	& (-.15,6.15)$\rightarrow\succsim^{3,U}$ & (-4.95,-1.05)$\rightarrow\succsim^{3,F}$	& (-1.75,-0.25)$\rightarrow\succsim^{3,F}$ \\
$S$3    & (-5.25,-0.75)$\rightarrow\succsim^{3,F}$	  & (-6.15,.15)$\rightarrow\succsim^{3,U}$	& (0,0)$\rightarrow\succsim^{3,T}$ & (-8.4,-3.6)$\rightarrow\succsim^{3,F}$ & (-6.4,-1.6)$\rightarrow\succsim^{3,F}$	 \\
$S$4    & (1.95,4.05)$\rightarrow\succsim^{3,T}$	  & (1.05,4.95)$\rightarrow\succsim^{3,T}$	& (3.6,8.4)$\rightarrow\succsim^{3,T}$	& (0,0)$\rightarrow\succsim^{3,T}$ & (0,0)$\rightarrow\succsim^{3,T}$\\
$S$5    & (-0.05,2.05)$\rightarrow\succsim^{3,U}$	  & (0.25,1.75)$\rightarrow\succsim^{3,T}$	& (1.6,6.4)$\rightarrow\succsim^{3,T}$	& (-3.2,-0.8)$\rightarrow\succsim^{3,F}$ & (0,0)$\rightarrow\succsim^{3,T}$\\
\hline
\end{tabular}
}
\end{table}


The central weight vector $\mathbf{w}^p$ and the vertex weight vectors belonging to sets $V(E^p_{(wp)}),\ p=1,2,3,$ are shown, together with the corresponding overall evaluations of the five students in each of the considered perspectives, in Tables \ref{Egalitarian}, \ref{Extreme} and \ref{Moderate}.

\begin{table}
\caption{Central and vertex weight vectors, and corresponding overall evaluations in the egalitarian perspective and value function aggregation}\label{Egalitarian}
\centering\footnotesize
\resizebox{1.0\textwidth}{!}{
\begin{tabular}{l|cccc|ccccc}
Weight  vector               & Mathematics    & Physics & Literature & Philosophy & S1 & S2 & S3 & S4 & S5 \\
\hline
$\mathbf{w}^1$  &  0.25	& 0.25	& 0.25	& 0.25 & 72.5 &	75	& 70	& 75	& 75\\
$\mathbf{\widehat{w}}^{1,1}$ 	& 0.2875	& 0.2875	& 0.2125	& 0.2125 & 74.38 &       75&            71.5 &          77.25 &         75.75\\
$\mathbf{\widehat{w}}^{1,2}$ 	& 0.2875	& 0.2125	& 0.2875	& 0.2125 & 71.38 &       75&            70.75 &          75 &         75\\
$\mathbf{\widehat{w}}^{1,3}$	& 0.2875	& 0.2125	& 0.2125	& 0.2875 & 72.88 &	74.25 &	72.25	& 75 &	75\\
$\mathbf{\widehat{w}}^{1,4}$	& 0.2125	& 0.2875	& 0.2875	& 0.2125 & 72.13	& 75.75	& 67.75	& 75	& 75\\
$\mathbf{\widehat{w}}^{1,5}$	& 0.2125	& 0.2875	& 0.2125	& 0.2875 & 73.63	& 75 &	69.25 &	75 &	75\\
$\mathbf{\widehat{w}}^{1,6}$	& 0.2125	& 0.2125	& 0.2875	& 0.2875 & 70.63 &	75	& 68.5 &	72.75	& 74.25\\
\hline
\end{tabular}
}
\end{table}


\begin{table}
\caption{Central and vertex  weight vectors, and corresponding overall evaluations in the extreme perspective and value function aggregation}\label{Extreme}
\centering\footnotesize
\resizebox{1.0\textwidth}{!}{
\begin{tabular}{l|cccc|ccccc}
Weight vector               & Mathematics    & Physics & Literature & Philosophy & S1 & S2 & S3 & S4 & S5 \\
\hline
$\mathbf{w}^2$  &  0.4 &	0.4	& 0.1 &	0.1 & 80	&	75	&	76	&	84	&	78\\
$\mathbf{\widehat{w}}^{2,    1            }$ &           0.46          &              0.37          &              0.085         &              0.085         &              80.3          &             74.55         &             78.4      &             84.9      &             78.3          \\            
$\mathbf{\widehat{w}}^{2,    2            }$ &           0.46          &              0.34          &              0.115         &              0.085         &              79.1          &             74.55         &             78.1          &             84            &             78            \\            
$\mathbf{\widehat{w}}^{2,    3            }$ &           0.46          &              0.34          &              0.085         &              0.115         &              79.7          &             74.25      &             78.7          &             84            &             78            \\            
$\mathbf{\widehat{w}}^{2,    4            }$ &           0.37          &              0.46          &              0.085         &              0.085         &              81.2          &             75.45         &             74.8          &             84.9      &             78.3          \\            
$\mathbf{\widehat{w}}^{2,    5            }$ &           0.34          &              0.46          &              0.115         &              0.085         &              80.3          &             75.75         &             73.3          &             84            &             78            \\            
$\mathbf{\widehat{w}}^{2,    6            }$ &           0.34          &              0.46          &              0.085         &              0.115         &              80.9      &             75.45         &             73.9          &             84            &             78            \\            
$\mathbf{\widehat{w}}^{2,    7            }$ &           0.43          &              0.34          &              0.115         &              0.115         &              78.8          &             74.55         &             77.2          &             83.1      &             77.7          \\            
$\mathbf{\widehat{w}}^{2,    8            }$ &           0.34          &              0.43          &              0.115         &              0.115         &              79.7          &             75.45         &             73.6      &             83.1      &             77.7          \\              
\hline
\end{tabular}
}
\end{table}

\begin{table}
\caption{Central and vertex  weight vectors, and corresponding overall evaluations in the moderate perspective and value function aggregation}\label{Moderate}
\centering\footnotesize
\resizebox{1.0\textwidth}{!}{
\begin{tabular}{l|cccc|ccccc}
Weight vector               & Mathematics    & Physics & Literature & Philosophy & S1 & S2 & S3 & S4 & S5 \\
\hline
$\mathbf{w}^3$  &  0.3 &	0.3	& 0.2 &	0.2 & 75	&	75	&	72	&	78	&	76\\
$\mathbf{\widehat{w}}^{3,    1            }$ &           0.345         &              0.315         &              0.17          &              0.17          &              76.35         &             74.85         &             73.8          &             79.8          &             76.6      \\            
$\mathbf{\widehat{w}}^{3,    2            }$ &           0.345         &              0.255         &              0.23          &              0.17          &              73.95         &             74.85         &             73.2          &             78      &             76            \\            
$\mathbf{\widehat{w}}^{3,    3            }$ &           0.345         &              0.255         &              0.17          &              0.23          &              75.15         &             74.25         &             74.4          &             78      &             76            \\            
$\mathbf{\widehat{w}}^{3,    4            }$ &           0.315         &              0.345         &              0.17          &              0.17          &              76.65         &             75.15         &             72.6          &             79.8          &             76.6      \\            
$\mathbf{\widehat{w}}^{3,    5            }$ &           0.255         &              0.345         &              0.23          &              0.17          &              74.85         &             75.75         &             69.6      &             78      &             76            \\            
$\mathbf{\widehat{w}}^{3,    6            }$ &           0.255         &              0.345         &              0.17          &              0.23          &              76.05         &             75.15         &             70.8          &             78      &             76            \\            
$\mathbf{\widehat{w}}^{3,    7            }$ &           0.285         &              0.255         &              0.23          &              0.23          &              73.35         &             74.85         &             71.4      &             76.2          &             76      \\            
$\mathbf{\widehat{w}}^{3,    8            }$ &           0.255         &              0.285         &              0.23          &              0.23          &              73.65         &             75.15         &             70.2          &             76.2          &             75.4      \\            
\hline
\end{tabular}
}
\end{table}

Applying Proposition 2, the overall evaluations of students shown in Tables \ref{Egalitarian}, \ref{Extreme} and \ref{Moderate} permit to deduce the preference relations $\succsim^{p,T}, \succsim^{p,F}$ and $\succsim^{p,U},\ p=1,2,3,$  which, obviously, are the same as presented in Tables  \ref{Ma_mi_egalitarian}, \ref{Ma_mi_extreme} and \ref{Ma_mi_moderate} for the corresponding perspectives.

Taking into account the preference relations $\succsim^{p,T}, \succsim^{p,F}$ and $\succsim^{p,U}$ in all considered perspectives $p=1,2,3$, one can deduce in turn the overall seven-valued preference relations between students, presented in Table \ref{pref_seven_valued}. 

\subsection{Explainability of seven-valued preferences}

\vspace{6pt}
The overall seven-valued preference relations presented to the dean may provoke the dean to raise some questions concerning \textbf{explainability}, and \textbf{robustness} of results, for example, ``why students $S2$ and $S3$ are in the `sometimes true' preference relation''? The methodology presented so far is traceable and permits to answer such questions in the following way. The overall preference relation between $S2$ and $S3$ is `sometimes true' because it is `true' in the egalitarian perspective (Table \ref{Ma_mi_egalitarian}), but `unknown' in the extreme (Table \ref{Ma_mi_extreme}) and moderate perspectives (Table \ref{Ma_mi_moderate}). To explain why this relation is `unknown' in the extreme perspective, let us come back to Table \ref{Extreme}, where overall evaluations of $S2$ and $S3$ are shown for central and vertex weight vectors. While $U(S2)\ge U(S3)$ for four vector weights where the weight of $Math$ is smaller than the weight of $Phys$, $U(S2) < U(S3)$ for five other weight vectors where the weight of $Math$ is at least as high as the weight of $Phys$. This means that in the extreme perspective, when $Math$ has a weight at least 0.4, and $Phys$ has a weight at most 0.4, the overall evaluation of $S2$ is worse than that of $S3$, and when the weight of $Math$ drops below 0.4 and the weight of $phys$ increases above 0.4, the overall evaluation of $S2$ is better than that of $S3$. For this reason, the relation between $S2$ and $S3$ is `unknown' in this perspective, i.e., $S2 \succsim^{2,U} S3$. In case of the moderate perspective, characterized in Table \ref{Moderate}, $U(S2)\ge U(S3)$ for all but one vector of weights. Indeed, $U(S2) < U(S3)$ only when the weight of $Lit$ drops to 0.17 and the weight of $Math$ increases to 0.345, which are the lowest and the highest values, respectively, in this perspective. In consequence, the relation between $S2$ and $S3$ is `unknown' also in this perspective, i.e., $S2 \succsim^{3,U} S3$. This explains why the overall preference relation between $S2$ and $S3$ is `sometimes true', i.e., $S2 \succsim^{sT} S3$.  

Another interesting question could be ``why students $S2$ and $S1$ are in the `fully contradictory' preference relation''? Remark that the preference relation between $S2$ and $S1$ is `true' in the egalitarian perspective, `false' in the extreme perspective, and `unknown' in the moderate perspective. The most striking difference between profiles of students $S2$ and $S1$ is in the grade of $Lit$, where $S2$ scored 80 and $S1$ scored 50. The overall advantage of $S2$ over $S1$ appears when the weights assigned to $Lit$ are equal or close to other weights, i.e., when they are not less than 0.2. This is the case of the egalitarian perspective (Table \ref{Egalitarian}) and the moderate perspective (Table \ref{Moderate}). When the weights of $Lit$ drop to 0.17 or less, at the expense of $Math$ and $Phys$, the overall advantage of $S1$ over $S2$ appears. This is the case of the extreme perspective (Table \ref{Extreme}) and the moderate perspective (Table \ref{Moderate}). This is why the overall preference relation between $S2$ and $S1$ is `fully contradictory', i.e., $S2 \succsim^{fK} S1$.

\begin{table}
\caption{Overall seven-valued preference relations between students for value function aggregation}\label{pref_seven_valued} 
\centering\footnotesize
\resizebox{0.45\textwidth}{!}{
\begin{tabular}{l|ccccc}
Student               & $S$1    & $S$2 & $S$3 & $S$4 & $S$5\\
\hline
$S$1    & $\succsim^{T}$	  & $\succsim^{fK}$	& $\succsim^{T}$	& $\succsim^{F}$ & $\succsim^{fK}$ \\
$S$2    & $\succsim^{fK}$	  & $\succsim^{T}$	& $\succsim^{sT}$ & $\succsim^{sF}$	& $\succsim^{sF}$ \\
$S$3    & $\succsim^{F}$	  & $\succsim^{sF}$	& $\succsim^{T}$ & $\succsim^{F}$ & $\succsim^{sF}$	 \\
$S$4    & $\succsim^{T}$	  & $\succsim^{sT}$	& $\succsim^{T}$	& $\succsim^{T}$ & $\succsim^{sT}$\\
$S$5    & $\succsim^{fK}$	  & $\succsim^{sT}$	& $\succsim^{sT}$	& $\succsim^{sF}$ & $\succsim^{T}$\\
\hline
\end{tabular}
}
\end{table}

\subsection{Seven-valued preferences and four-valued logic}

Continuing the analysis of the obtained seven-valued preference relations, it is interesting to note that some of them could be aggregated to form a less fine four-valued preference structure in the following manner: for all pairs of students $S$ and $S'$, 
\begin{itemize}
	\item there is true preference of $S$ over $S'$, denoted by $S\succsim_4^{T}S'$, if $S\succsim^{T}S'$ or $S\succsim^{sT}S'$,
	\item there is unknown preference between $S$ and $S'$, denoted by $S\succsim_4^{U}S'$, if $S\succsim^{U}S'$,
	\item there is contradictory preference between $S$ and $S'$, denoted by $S\succsim_4^{K}S'$, if $S\succsim^{K}S'$ or $S\succsim^{fK}S'$,
	\item there is false preference of $S$ over $S'$, denoted by $S\succsim_4^{F}S'$, if $S\succsim^{sF}S'$ or $S\succsim^{F}S'$.
\end{itemize}
Note that, in the spirit of Belnap's four-valued logic \cite{Belnap76,Belnap77}, the above four-valued preference structure can be described as follows.  
There is an argument in favor of the preference of $S$ over $S'$  if $S\succsim^{p,T}S$ for some perspective $p \in \{1,2,3\}$, while there is an argument against the preference of $S$ over $S'$  if $S\succsim^{p,F}S$ for some perspective $p \in \{1,2,3\}$. Following this logic, for all students $S$ and $S'$, we have:
\begin{itemize}
	\item there is true preference of $S$ over $S'$ if there is some argument in favor and there is no argument against, that is, $S\succsim_4^{T}S'$, if $S\succsim^{pT}S'$ for some $p \in \{1,2,3,\}$ and there is no $p\in\{1,2,3\}$ for which $S\succsim^{pF}S'$,
	\item there is unknown preference between $S$ and $S'$ if there is no argument in favor and there is no argument against, that is, $S\succsim_4^{U}S'$, if there is no $p \in \{1,2,3,\}$ for which $S\succsim^{pT}S'$  and there is no $p\in\{1,2,3\}$ for which $S\succsim^{pF}S'$,
	\item there is contradictory preference between $S$ and $S'$ if there is some argument in favor and there is some argument against, that is,  $S\succsim_4^{K}S'$, if there is some $p\in\{1,2,3\}$ for which $S\succsim^{pF}S'$ and $S\succsim^{pT}S'$ for some other $p \in \{1,2,3\}$.	
	\item there is false preference of $S$ over $S'$ if there is some argument against and there is no argument in favor, that is, $S\succsim_4^{F}S'$, if $S\succsim^{pF}S'$ for some $p \in \{1,2,3,\}$ and there is no $p\in\{1,2,3\}$ for which $S\succsim^{pT}S'$.
\end{itemize}

\vspace{6pt}

\subsection{Utilization of the seven-valued preference relations in view of making a ranking recommendation}

The dean's ultimate goal is to derive the overall ranking of students from the seven-valued preference relations among them. To achieve this, a global score $V^G(S)$ is calculated for each student $S$, based on  how $S$ compares to all other students, $S'$, using the seven-valued preference relations. In particular, in the global score of $S$, a specific gain or loss value,  $v(S \succsim^{H} S') \geq 0$, is assigned to each of the seven possible preference relations between $S$ and $S'$, i.e., $S \succsim^{H} S',\ H \in\{T,sT,U,K,fK,sF,F\}$. Similarly, a specific gain or loss value,  $v(S' \succsim^{H} S) \geq 0$, is assigned to each of the seven possible preference relations between $S'$ and $S$, i.e., $S' \succsim^{H} S,\ H\in\{T,sT,U,K,fK,sF,F\}$. The values assigned to the gains or losses, $v(S \succsim^{H} S')$ and $v(S' \succsim^{H} S)$, have to respect the following conditions:
\begin{itemize}
	\item the gain in the global score of student $S$ in case of `true' preference $S \succsim^{T} S'$ and `sometimes true' preference $S \succsim^{sT} S'$ is non-negative, i.e., \linebreak $v(S \succsim^{T} S') \geq 0$ and $v(S \succsim^{sT} S') \geq 0$,
	\item the loss in the global score of student $S$ in case of `false' preference $S \succsim^{F} S'$ and `sometimes false' preference $S \succsim^{sF} S'$ is non-negative, i.e., \linebreak $v(S \succsim^{F} S') \geq 0$ and $v(S \succsim^{sF} S') \geq 0$,
	\item the loss in the global score of student $S$ in case of `true' inverse preference $S' \succsim^{T} S$ and `sometimes true' inverse preference $S' \succsim^{sT} S$ is non-negative, i.e., $v(S' \succsim^{T} S) \geq 0$ and $v(S' \succsim^{sT} S) \geq 0$,
	\item the gain in the global score of student $S$ in case of `false' inverse preference $S' \succsim^{F} S$ and `sometimes false' inverse preference $S' \succsim^{sF} S$ is non-negative, i.e., $v(S' \succsim^{F} S) \geq 0$ and $v(S' \succsim^{sF} S) \geq 0$,
	\item the gain in the global score of student $S$ in case of `true' preference $S \succsim^{T} S'$ cannot have a value smaller than the gain of `sometimes true' preference $S \succsim^{sT} S'$, so that $v(S \succsim^{T} S') \geq v(S \succsim^{sT} S')$,
	\item the loss in the global score of student $S$ in case of `false' preference \linebreak $S \succsim^{F} S'$ cannot have a value smaller than the loss of `sometimes false' preference $S \succsim^{sF} S'$, so that $v(S \succsim^{F} S') \geq v(S \succsim^{sF} S')$,
	\item the loss in the global score of student $S$ in case of `true' inverse preference \linebreak $S' \succsim^{T} S$ cannot have a value smaller than the loss of `sometimes true' inverse preference $S' \succsim^{sT} S$, so that $v(S' \succsim^{T} S) \geq v(S' \succsim^{sT} S)$,
	\item the gain in the global score of student $S$ in case of `false' inverse preference $S' \succsim^{F} S$ cannot have a value smaller than the gain of `sometimes false' inverse preference $S' \succsim^{sF} S$, so that $v(S' \succsim^{F} S) \geq v(S' \succsim^{sF} S)$,
	\item a null value adds to the global score of student $S$ in case of `unknown', `contradictory' and `fully contradictory' preference and inverse preference, i.e., 	$v(S \succsim^{H} S')=v(S' \succsim^{H} S)=0,\ H \in \{U, K, fK\}$.	
\end{itemize}

Consequently, the global score of student $S$ is calculated as:
$$
\displaystyle{V^G(S)=\sum_{\forall S'\neq S} \sum_{H\in\{T,sT\}} v(S\succsim^{H} S')-\sum_{\forall S'\neq S}\sum_{H\in\{sF,F\}} v(S\succsim^{H} S')} 
$$ 
$$
\displaystyle{ - \sum_{\forall S'\neq S}\sum_{H\in\{T,sT\}} v(S'\succsim^{H} S)+\sum_{\forall S'\neq S} \sum_{H\in\{sF,F\}}v(S'\succsim^{H} S)}.
$$

Initially, the dean used the following `basic' convention to assign values to gains and losses $v(S \succsim^{H} S'),\ v(S' \succsim^{H} S),\ H\in\{T,sT,U,K,fK,sF,F\}$:
\begin{itemize}
	\item $v(S \succsim^{T} S')=v(S' \succsim^{F} S)=1,$
	\item $v(S \succsim^{sT} S')=v(S' \succsim^{sF} S)=0.5,$
	\item $v(S \succsim^{U} S')=v(S \succsim^{K} S')=v(S \succsim^{fK} S')=0,$ \newline as well as $v(S' \succsim^{U} S)=v(S' \succsim^{K} S)=v(S' \succsim^{fK} S)=0,$
	\item $v(S \succsim^{sF} S')=v(S' \succsim^{sT} S)=0.5,$
	\item $v(S \succsim^{F} S')=v(S' \succsim^{T} S)=1.$
\end{itemize}
In doing so, the global scores obtained by students is as follows: 
$$V^G(S1)=0,\ V^G(S2)=-1,\ V^G(S3)=-6,\ V^G(S4)=6,\ V^G(S5)=1.$$ 
Thus, the ranking of students according to the above way of utilization of the overall seven-valued preference relations is: $S4 \rightarrow S5 \rightarrow S1 \rightarrow S2 \rightarrow S3$.
\vspace{6pt}

Later, to determine values of gains and losses $v(S \succsim^{H} S'),\ v(S' \succsim^{H} S)$, $H\in\{T,sT,U,K,fK,sF,F\}$ the dean decided to use the `deck of cards' method, assuming that $v(S \succsim^{T} S')=v(S' \succsim^{F} S)$, $v(S \succsim^{sT} S')=v(S' \succsim^{sF} S)$, \linebreak $v(S \succsim^{sF} S')=v(S' \succsim^{sT} S)$, and $v(S \succsim^{F} S')=v(S' \succsim^{T} S)$. Moreover, a null value is assigned again to  `unknown', `contradictory' and `fully contradictory' preference and inverse preference, i.e., 	$v(S \succsim^{H} S')=v(S' \succsim^{H} S)=0,$ $H \in \{U, K, fK\}$. 

The `deck of cards' method proceeds in the following steps:  

\begin{itemize}
	\item Step 1: the dean places a number of cards, $e(F,sF)$, between $\succsim^{F}$ and $\succsim^{sF}$, representing the difference in value between $v(S \succsim^{F} S')$ and $v(S \succsim^{sF} S')$; similarly, the dean places a number of cards, $e(sF,\{U,K,fK\})$, between $F$ and $\{U,K,fK\}$, a number of cards, $e(\{U,K,fK\},sT)$, between $\{U,K,fK\}$ and $sT$, and a number of cards, $e(sT,T)$, between $sT$ and $T$;
	\item Step 2: the following non-normalized values $\nu(S \succsim^{H} S')$,   $H\in\{T,sT,U,K,\linebreak fK,sF,F\}$, are assigned:
	\begin{itemize}
		\item $\nu(S \succsim^{U} S')=\nu(S \succsim^{K} S')=\nu(S \succsim^{fK} S')=0$,
		\item $\nu(S \succsim^{sT} S')=e(\{U,K,fK\},sT)+1$,
		\item $\nu(S \succsim^{T} S')=\nu(S \succsim^{sT} S')+e(sT,T)+1$,
		\item $\nu(S \succsim^{sF} S')=e(sF,\{U,K,fK\})+1$,
		\item $\nu(S \succsim^{F} S')=\nu(S \succsim^{sF} S')+e(F,sF)+1$;	
	\end{itemize}
	\item Step 3: the values of gains and losses, $v(S \succsim^{H} S')$, $H\in\{T,sT,U,K,fK,sF,F\}$, are obtained by dividing the non-normalized values $\nu(S \succsim^{H} S')$ by \linebreak $max\big\{\nu(S \succsim^{T} S'), \nu(S \succsim^{F} S')\big\}$, that is, 
	$$v(S \succsim^{H} S')=\frac{\nu(S \succsim^{H} S')}{max\big\{\nu(S \succsim^{T} S'), \nu(S \succsim^{F} S')\big\}}.$$
\end{itemize}

In particular, the dean places the following number of cards:
\begin{itemize}
	\item $e(F,sF)=6$ cards between $\succsim^{F}$ and $\succsim^{sF}$,
	\item $e(sF,\{U,K,fK\})=5$ cards between $\succsim^{sF}$ and $\succsim^{H},\ H\in\{U,K,fK\}$,
	\item $e(\{U,K,fK\}),sT)=3$ cards between $\succsim^{H}, H\in\{U,K,fK\}$, and $\succsim^{sT}$,
	\item $e(sT,T)=2$ cards between $\succsim^{sT}$ and $\succsim^{T}$.
\end{itemize}

In doing so, the `deck-of-cards' method yields the following non-normalized values $\nu(S \succsim^{H} S')$, $H\in\{T,sT,U,K,fK,sF,F\}$:
\begin{itemize}
	\item $\nu(S \succsim^{U} S')=\nu(S \succsim^{K} S')=\nu(S \succsim^{fK} S')=0$,
	\item $\nu(S \succsim^{sT} S')=4$,
	\item $\nu(S \succsim^{T} S')=7$,
	\item $\nu(S \succsim^{sF} S')=6$,
	\item $\nu(S \succsim^{F} S')=13$.	
\end{itemize}

By dividing the above-mentioned non-normalized values $\nu(S \succsim^{H} S')$, \linebreak $H\in\{T,sT,U,K,fK,sF,F\}$ by $max\big\{\nu(S \succsim^{T} S'), \nu(S \succsim^{F} S')\big\}=max\{4,13\}=13$, we get the following values for the gains or losses $v(S \succsim^{H} S'), H\in\{T,sT,U,K, \linebreak fK,sF,F\}$:
\begin{itemize}
	\item $v(S \succsim^{T} S')=0.54,$
	\item $v(S \succsim^{sT} S')=0.31,$
	\item $v(S \succsim^{U} S')=v(S \succsim^{K} S')=v(S \succsim^{fK} S')=0,$
	\item $v(S \succsim^{sF} S')=0.46,$
	\item $v(S \succsim^{F} S')=1.$
\end{itemize}
In consequence, the global scores obtained by students are the following: 
$$V^G(S1)=0, \ V^G(S2)=-0.77, \ V^G(S3)=-4.62, \ V^G(S4)=4.62, \ V^G(S5)=0.77.$$ 
Thus, the ranking of students is the same as before: $S4 \rightarrow S5 \rightarrow S1 \rightarrow S2 \rightarrow S3$.

\subsection{Construction of the seven-valued preference relations with outranking aggregation}

\vspace{6pt}
Let us change now the weighted sum value function to an outranking function used in ELECTRE-like methods. Suppose that the dean adopts the same weight-vectors as shown in Tables \ref{Egalitarian}, \ref{Extreme} and \ref{Moderate}, however, in this case, the central weights were determined using a procedure coherent with the meaning of weights in ELECTRE-like methods, i.e., not as trade-off weights but as relative strengths in a voting procedure. The `deck of the cards' method described in \cite{Figueira_Roy} is appropriate for this task.

For all pairs of students, $S$ and $S'$, for $\mathcal{S}$ being the set of subjects, and for all weight vectors $\mathbf{\widetilde{w}}^p=[\widetilde{w}^p_{Math},\widetilde{w}^p_{Phys},\widetilde{w}^p_{Lit},\widetilde{w}^p_{Phil}]$ from set $E^p_{ (wp)}$,
$S$ outranks $S'$, denoted by $S \succsim(\mathbf{\widetilde{w}}^p) S'$, if 
\[
\displaystyle{C(S \succsim(\mathbf{\widetilde{w}}^p) S')=\sum_{s_j \in \mathcal{S}:\ g_{s_j}(S) \geqslant g_{s_j}(S')-q}\widetilde{w}^p_{s_j} \geqslant k}
\]
with a chosen indifference threshold $q \geqslant 0$ and an opportune concordance level $k\in(0.5,1]$. 

Taking into account  the outranking relations $\succsim(\mathbf{\widetilde{w}}^p),\ \widetilde{w}^p \in E^p_{ (wp)}, p=1,2,3$, one can conclude that the proposition ``student $S$ is at least as good as student $S'$'' is:
\begin{itemize}
	\item  true, and denoted by $S \succsim^{p,T} S'$, if $S \succsim(\mathbf{\widetilde{w}}^p) S'$ for all $\mathbf{\widetilde{w}}^p \in E^p_{ (wp) }$,
	\item false, and denoted by $S \succsim^{p,F} S'$, if \textit{not} $S \succsim(\mathbf{\widetilde{w}}^p) S'$ for all $\mathbf{\widetilde{w}}^p \in E^p_{ (wp) }$,
	\item unknown, and denoted by $S \succsim^{p,U} S'$, if $S \succsim(\mathbf{\widetilde{w}}^p) S'$ for some $\mathbf{\widetilde{w}}^p \in E^p_{ (wp) }$ and \textit{not} $S \succsim(\mathbf{\widetilde{w}}^p) S'$ for some other $\mathbf{\widetilde{w}}^p \in E^p_{ (wp) }$.
\end{itemize}

The outranking relations $\succsim^{p,T}, \succsim^{p,F}$ and $\succsim^{p,U}$ can be computed on the basis of the  following Proposition 3 and Proposition 4,  analogous to 
Proposition 1 and Proposition 2 for value function aggregation.
\vspace{6pt}

\textbf{Proposition 3.} \textit{For all pairs of students, $S$ and $S'$, and constraints $E^p_{ (wp)}$ on perturbed weight vectors in one perspective $p \in \{1,2,3\}$, it holds that:
\begin{itemize}
	\item $S \succsim^{p,T} S'$ if and only if  $m_{out}^p(S,S') \geqslant 0$,
	\item $S \succsim^{p,F} S'$ if and only if  $M_{out}^p(S,S') < 0$,
	\item $S \succsim^{p,U} S'$ if and only if $m_{out}^p(S,S') < 0 \leqslant M_{out}^p(S,S')$,
	\end{itemize}
	with
	\begin{itemize}
		\item $m_{out}^p(S,S')=min[C(S \succsim(\mathbf{\widetilde{w}}^p) S')-k]$ subject to $E^p_{ (wp)}$,
		\item $M_{out}^p(S,S')=max [C(S \succsim(\mathbf{\widetilde{w}}^p) S')-k]$ subject to $E^p_{ (wp)}$.
	\end{itemize}}

\vspace{6pt}

\textbf{Proposition 4.} \textit{For all pairs of students, $S$ and $S'$, and constraints $E^p_{ (wp)}$ on perturbed weight vectors in one perspective $p \in \{1,2,3\}$,  it holds that:
	\begin{itemize}
		\item $S \succsim^{p,T} S'$ if and only if  $C(S \succsim(\mathbf{\widetilde{w}}^p) S') \geqslant k$ for all $\mathbf{\widetilde{w}}^p \in V(E^p_{ (wp)})$,
		\item $S \succsim^{p,F} S'$ if and only if  $C(S \succsim(\mathbf{\widetilde{w}}^p) S') < k$ for all $\mathbf{\widetilde{w}}^p \in V(E^p_{ (wp)})$,
		\item $S \succsim^{p,U} S'$ if and only if $C(S \succsim(\mathbf{\widetilde{w}}^p) S') \geqslant k$ for some $\mathbf{\widetilde{w}}^p \in V(E^p_{ (wp)})$ and $C(S \succsim(\mathbf{\widetilde{w}}^p) S') < k$ for some other $\mathbf{\widetilde{w}}^p \in V(E^p_{ (wp)})$.
	\end{itemize}
}
The proofs of Propositions 3 and 4 are analogous to those of Proposition 1 and 2.

Suppose that the dean set the indifference threshold at $q=1$ and concordance level at $k=0.65$, obtaining the true, false, and unknown outranking relations, $\succsim^{p,T},\ \succsim^{p,F}$ and $\succsim^{p,U},\ p=1,2,3$,  presented in Tables  \ref{out_Egalitarian}, \ref{out_Extreme}, and \ref{out_Moderate}, for the corresponding perspectives.

\begin{table}
\caption{Outranking relations between students in the egalitarian perspective:\linebreak $\succsim^{1,T}$, $\succsim^{1,F}$, and $\succsim^{1,U}$} \label{out_Egalitarian}
\centering\footnotesize
\resizebox{0.45\textwidth}{!}{
\begin{tabular}{l|ccccc}
Student               & $S$1    & $S$2 & $S$3 & $S$4 & $S$5\\
\hline
$S$1    & $\succsim^{1,T}$	  & $\succsim^{1,T}$	& $\succsim^{1,T}$	& $\succsim^{1,F}$ & $\succsim^{1,T}$ \\
$S$2    & $\succsim^{1,F}$	  & $\succsim^{1,T}$	& $\succsim^{1,T}$ & $\succsim^{1,F}$	& $\succsim^{1,T}$ \\
$S$3    & $\succsim^{1,T}$	  & $\succsim^{1,F}$	& $\succsim^{1,T}$ & $\succsim^{1,F}$ & $\succsim^{1,F}$	 \\
$S$4    & $\succsim^{1,T}$	  & $\succsim^{1,F}$	& $\succsim^{1,F}$	& $\succsim^{1,T}$ & $\succsim^{1,F}$\\
$S$5    & $\succsim^{1,T}$	  & $\succsim^{1,T}$	& $\succsim^{1,T}$	& $\succsim^{1,F}$ & $\succsim^{1,T}$\\
\hline
\end{tabular}
}
\end{table}


\begin{table}
\caption{Outranking relations between students in the extreme perspective:\linebreak $\succsim^{2,T}$, $\succsim^{2,F}$, and $\succsim^{2,U}$} \label{out_Extreme}
\centering\footnotesize
\resizebox{0.45\textwidth}{!}{
\begin{tabular}{l|ccccc}
Student               & $S$1    & $S$2 & $S$3 & $S$4 & $S$5\\
\hline
$S$1    & $\succsim^{2,T}$	  & $\succsim^{2,T}$	& $\succsim^{2,U}$	& $\succsim^{2,F}$ & $\succsim^{2,T}$ \\
$S$2    & $\succsim^{2,F}$	  & $\succsim^{2,T}$	& $\succsim^{2,U}$ & $\succsim^{2,F}$	& $\succsim^{2,U}$ \\
$S$3    & $\succsim^{2,U}$	  & $\succsim^{2,F}$	& $\succsim^{2,T}$ & $\succsim^{2,F}$ & $\succsim^{2,F}$	 \\
$S$4    & $\succsim^{2,T}$	  & $\succsim^{2,T}$	& $\succsim^{2,F}$	& $\succsim^{2,T}$ & $\succsim^{2,T}$\\
$S$5    & $\succsim^{2,U}$	  & $\succsim^{2,T}$	& $\succsim^{2,U}$	& $\succsim^{2,F}$ & $\succsim^{2,T}$\\
\hline
\end{tabular}
}
\end{table}


\begin{table}
\caption{Outranking relations between students in the moderate perspective:\linebreak $\succsim^{3,T}, \succsim^{3,F}$ and $\succsim^{3,U}$} \label{out_Moderate}
\centering\footnotesize
\resizebox{0.45\textwidth}{!}{
\begin{tabular}{l|ccccc}
Student               & $S$1    & $S$2 & $S$3 & $S$4 & $S$5\\
\hline
$S$1    & $\succsim^{3,T}$	  & $\succsim^{3,T}$	& $\succsim^{3,T}$	& $\succsim^{3,F}$ & $\succsim^{3,T}$ \\
$S$2    & $\succsim^{3,F}$	  & $\succsim^{3,T}$	& $\succsim^{3,T}$ & $\succsim^{3,F}$	& $\succsim^{3,T}$ \\
$S$3    & $\succsim^{3,T}$	  & $\succsim^{3,F}$	& $\succsim^{3,T}$ & $\succsim^{3,F}$ & $\succsim^{3,F}$	 \\
$S$4    & $\succsim^{3,T}$	  & $\succsim^{3,U}$	& $\succsim^{3,F}$	& $\succsim^{3,T}$ & $\succsim^{3,U}$\\
$S$5    & $\succsim^{3,T}$	  & $\succsim^{3,T}$	& $\succsim^{3,T}$	& $\succsim^{3,F}$ & $\succsim^{3,T}$\\
\hline
\end{tabular}
}
\end{table}


Taking into account the preference relations $\succsim^{p,T},\ \succsim^{p,F}$, and $\succsim^{p,U}$, in all considered perspectives $p=1,2,3$, one can deduce in turn the overall seven-valued preference relations between students, presented in Table \ref{pref_seven-valued}. 

\begin{table}
\caption{Overall seven-valued preference relations between students for outranking aggregation} \label{pref_seven-valued}
\centering\footnotesize
\resizebox{0.42\textwidth}{!}{
\begin{tabular}{l|ccccc}
Student               & $S$1    & $S$2 & $S$3 & $S$4 & $S$5\\
\hline
$S$1    & $\succsim^{T}$	  & $\succsim^{T}$	& $\succsim^{sT}$	& $\succsim^{F}$ & $\succsim^{T}$ \\
$S$2    & $\succsim^{F}$	  & $\succsim^{T}$	& $\succsim^{sT}$ & $\succsim^{F}$	& $\succsim^{sT}$ \\
$S$3    & $\succsim^{T}$	  & $\succsim^{F}$	& $\succsim^{T}$ & $\succsim^{F}$ & $\succsim^{F}$	 \\
$S$4    & $\succsim^{T}$	  & $\succsim^{fK}$	& $\succsim^{F}$	& $\succsim^{T}$ & $\succsim^{fK}$\\
$S$5    & $\succsim^{T}$	  & $\succsim^{T}$	& $\succsim^{sT}$	& $\succsim^{F}$ & $\succsim^{T}$\\
\hline
\end{tabular}
}
\end{table}


Applying the ``basic'' values of gains and losses $v(S \succsim^{H} S')$, $v(S' \succsim^{H} S),\ H\in\{T,sT,U,K,fK,sF,F\}$, to the seven-valued outranking shown in Table \ref{pref_seven-valued}, the five students were assigned the following global scores:
$$V^G(S1)=-0.5, \ V^G(S2)=-2, \ V^G(S3)=-2.5, \ V^G(S4)=4, \ V^G(S5)=1,$$ 
resulting in the same ranking as above, that is, $S4 \rightarrow S5 \rightarrow S1 \rightarrow S2 \rightarrow S3$.

Using the `deck-of-cards' method for finding values of gains and losses, in the same way as in the case of value function aggregation, the dean obtained the following global scores:
$$V^G(S1)=-0.23, \ V^G(S2)=-1.46, \ V^G(S3)=-2.38, \ V^G(S4)=3.54, \ V^G(S5)=0.54,$$ 
resulting in the same ranking as above.

\subsection{Addressing robustness concerns through Stochastic Multicriteria Acceptability Analysis}
		
\vspace{6pt}
To avoid bias in the seven-valued preference relations resulting from overall evaluations by value functions with weight vectors located only at the vertices of $E^p_{(wp)}$, the dean considered the probability $Pr(S \succsim S')$ of student $S$ being preferred over student $S'$. These probabilities, called ``pairwise winning indices'', were obtained using SMAA (Stochastic Multicriteria Acceptability Analysis) \cite{SMAA,SMAA_2} with a uniform probability distribution in the space of feasible weights, and, more precisely, using the `hit-and-run' algorithm in the simplex $E^p_{(wp)}$ with a random sampling of 100,000 weight vectors for each perspective $p=1,2,3$. The results obtained for the three perspectives are shown in Tables \ref{SMAA_Egalitarian}, \ref{SMAA_Extreme}, \ref{SMAA_Moderate}, respectively.   		

\begin{table}
\caption{Pairwise winning indices of students in rows over students in columns in the egalitarian perspective and value function aggregation} \label{SMAA_Egalitarian}
\centering\footnotesize
\resizebox{0.4\textwidth}{!}{
\begin{tabular}{l|ccccc}
Student               & $S$1    & $S$2 & $S$3 & $S$4 & $S$5\\
\hline
$S$1    & 1	  & 0	   & 1	& 0 & 0 \\
$S$2    & 1	  & 1 	   & 1  & 0.51	& 0.51 \\
$S$3    & 0	  & 0	   & 1  & 0 & 0	 \\
$S$4    & 1	  & 0.49   & 1	& 1 & 0.5\\
$S$5    & 1	  & 0.49   & 1	& 0.5 & 1\\
\hline
\end{tabular}
}
\end{table}


\begin{table}
\caption{Pairwise winning indices of students in rows over students in columns in  the extreme perspective and value function aggregation} \label{SMAA_Extreme}
\centering\footnotesize
\resizebox{0.4\textwidth}{!}{
\begin{tabular}{l|ccccc}
Student               & $S$1    & $S$2 & $S$3 & $S$4 & $S$5\\
\hline
$S$1    & 1	  & 1	& 1	& 0 & 1 \\
$S$2    & 0	  & 1	& 0.35 & 0	& 0 \\
$S$3    & 0	  & 0.65	& 1 & 0 & 0.06	 \\
$S$4    & 1	  & 1	& 1	& 1 & 1\\
$S$5    & 0	  & 1	& 0.94	& 0 & 1\\
\hline
\end{tabular}
}
\end{table}

\begin{table}
\caption{Pairwise winning indices of students in rows over students in columns in the moderate perspective and value function aggregation} \label{SMAA_Moderate}
\centering\footnotesize
\resizebox{0.4\textwidth}{!}{
\begin{tabular}{l|ccccc}
Student               & $S$1    & $S$2 & $S$3 & $S$4 & $S$5\\
\hline
$S$1    & 1	  & 0.5	& 1	& 0 & 0.01 \\
$S$2    & 0.5	  & 1	& 1 & 0	& 0 \\
$S$3    & 0	  & 0	& 1 & 0 & 0	 \\
$S$4    & 1	  & 1	& 1	& 1 & 1\\
$S$5    & 0.99	  & 1	& 1	& 0 & 1\\
\hline
\end{tabular}
}
\end{table}


Taking into account the pairwise winning indices from Tables \ref{SMAA_Egalitarian}, \ref{SMAA_Extreme}, and \ref{SMAA_Moderate}, and setting a threshold of $t \in (0.5,1]$ on these probabilities, the true, false, and unknown preference relations, $\succsim^{p,T},\ \succsim^{p,F}$ and $\succsim^{p,U},\ p\in\{1,2,3\}$ are obtained:
\begin{itemize}
	\item $S \succsim^{p,T} S'$, if $Pr(S \succsim S') \geqslant t$,
	\item $S \succsim^{p,F} S'$, if $Pr(S \succsim S') \leqslant 1-t$,
	\item $S \succsim^{p,U} S'$, if $1- t < Pr(S \succsim S') < t$.
\end{itemize}

For example, setting $t=0.85$, the preference relations in Tables \ref{Ma_mi_egalitarian} and \ref{Ma_mi_extreme} remain the same, while the preferences in Table \ref{Ma_mi_moderate} have to be ``corrected'', as shown in Table \ref{pref_Moderate_}, where the original values are put in parentheses when modified.

\begin{table}
\caption{Preference relations between students based on pairwise winning indices in  the moderate perspective and value function aggregation: $\succsim^{3,T},\ \succsim^{3,F}$, and $\succsim^{3,U}$} \label{pref_Moderate_}
\centering\footnotesize
\resizebox{0.8\textwidth}{!}{
\begin{tabular}{l|c|c|c|c|c}
Student               & $S$1    & $S$2 & $S$3 & $S$4 & $S$5\\
\hline
$S$1    & $\succsim^{3,T}$	  & $\succsim^{3,U}$	& $\succsim^{3,T}$	& $\succsim^{3,F}$ & $\succsim^{3,F}$($\succsim^{3,U}$) \\
$S$2    & $\succsim^{3,U}$	  & $\succsim^{3,T}$	& $\succsim^{3,T}$($\succsim^{3,U}$) & $\succsim^{3,F}$	& $\succsim^{3,F}$ \\
$S$3    & $\succsim^{3,F}$	  & $\succsim^{3,F}$($\succsim^{3,U}$)	& $\succsim^{3,T}$ & $\succsim^{3,F}$ & $\succsim^{3,F}$	 \\
$S$4    & $\succsim^{3,T}$	  & $\succsim^{3,T}$	& $\succsim^{3,T}$	& $\succsim^{3,T}$ & $\succsim^{3,T}$\\
$S$5    & $\succsim^{3,T}$($\succsim^{3,U}$)  & $\succsim^{3,T}$	& $\succsim^{3,T}$	& $\succsim^{3,F}$ & $\succsim^{3,T}$\\
\hline
\end{tabular}
}
\end{table}


Applying the ``corrections'' resulting from consideration of pairwise winning indices in the value function approach, the overall seven-valued preference relations between students shown in Table \ref{pref_seven-valued} remained unchanged, except for the preference relation between students $S1$ and $S5$. Specifically, now $S1 \succsim^K S5$ and $S5 \succsim^K S1$, whereas previously it was $S1 \succsim^{fK} S5$ and $S5 \succsim^{fK} S1$. The global netflow scores and the final ranking of students remained the same.  

\vspace{6pt}
Continuing the analysis, the dean also wished to verify the stability of the outranking relations from three perspectives using the same probabilistic approach adopted for the value function-based relations. To this end, the probabilities that one student outranks another, called ``pairwise winning indices'' as before, using a randomly selected feasible weight vector from $E^p_{(wp)}$ were computed for the three perspectives, as shown in Tables \ref{SMAA_Egalitarian_out}, \ref{SMAA_Extreme_out}, \ref{SMAA_Moderate_out}, respectively.



\begin{table}
	\caption{Pairwise winning indices of students in rows over students in columns in the egalitarian perspective and outranking aggregation} \label{SMAA_Egalitarian_out}
	\centering\footnotesize
	\resizebox{0.35\textwidth}{!}{
		\begin{tabular}{l|ccccc}
			Student               & $S$1    & $S$2 & $S$3 & $S$4 & $S$5\\
			\hline
			$S$1    & 1	  & 1	& 1	& 0 & 1 \\
			$S$2    & 0	  & 1	& 1 & 0	& 1 \\
			$S$3    & 1	  & 0	& 1 & 0 & 0	 \\
			$S$4    & 1	  & 0	& 0	& 1 & 0\\
			$S$5    & 1	  & 1	& 1	& 0 & 1\\
			\hline
		\end{tabular}
	}
\end{table}
%

\begin{table}
	\caption{Pairwise winning indices of students in rows over students in columns in the extreme perspective and outranking aggregation} \label{SMAA_Extreme_out}
	\centering\footnotesize
	\resizebox{0.4\textwidth}{!}{
		\begin{tabular}{l|ccccc}
			Student               & $S$1    & $S$2 & $S$3 & $S$4 & $S$5\\
			\hline
			$S$1    & 1	  & 1	& 0.08	& 0 & 1 \\
			$S$2    & 0	  & 1	& 0.08 & 0	& 0.08 \\
			$S$3    & 0.07	  & 0	& 1 & 0 & 0	 \\
			$S$4    & 1	  & 1	& 0	& 1 & 1\\
			$S$5    & 0.07	  & 1	& 0.08	& 0 & 1\\
			\hline
		\end{tabular}
	}
\end{table}

\begin{table}
	\caption{Pairwise winning indices of students in rows over students in columns in the moderate perspective and outranking aggregation} \label{SMAA_Moderate_out}
	\centering\footnotesize
	\resizebox{0.4\textwidth}{!}{
		\begin{tabular}{l|ccccc}
			Student               & $S$1    & $S$2 & $S$3 & $S$4 & $S$5\\
			\hline
			$S$1    & 1	  & 1	& 1	& 0 & 1 \\
			$S$2    & 0	  & 1	& 1 & 0	& 1 \\
			$S$3    & 1	  & 0	& 1 & 0 & 0	 \\
			$S$4    & 1	  & 0.02	& 1	& 1 & 0.02\\
			$S$5    & 1	  & 1	& 1	& 0 & 1\\
			\hline
		\end{tabular}
	}
\end{table}


Taking into account the pairwise winning indices from Tables \ref{SMAA_Egalitarian_out}, \ref{SMAA_Extreme_out}, and \ref{SMAA_Moderate_out}, and setting a threshold of  $t=0.85$ on these probabilities, the outranking relations remained unchanged in the egalitarian perspective, however, they changed in the extreme and moderate perspectives, as shown in Tables \ref{out_Extreme_corr} and \ref{out_Moderate_out}, where the original values are put in parentheses when modified. 

\begin{table}
	\caption{Outranking relations between students based on pairwise winning indices in the extreme perspective and outranking aggregation: $\succsim^{2,T}$, $\succsim^{2,F}$, and $\succsim^{2,U}$} \label{out_Extreme_corr}
	\centering\footnotesize
	\resizebox{0.75\textwidth}{!}{
		\begin{tabular}{l|c|c|c|c|c}
			Student               & $S$1    & $S$2 & $S$3 & $S$4 & $S$5\\
			\hline
			$S$1    & $\succsim^{2,T}$	  & $\succsim^{2,T}$	& $\succsim^{2,F}$($\succsim^{2,U}$)	& $\succsim^{2,F}$ & $\succsim^{2,T}$ \\
			$S$2    & $\succsim^{2,F}$	  & $\succsim^{2,T}$	& $\succsim^{2,F}$($\succsim^{2,U}$)	 & $\succsim^{2,F}$	& $\succsim^{2,F}$($\succsim^{2,U}$)	 \\
			$S$3    & $\succsim^{2,F}$($\succsim^{2,U}$)		  & $\succsim^{2,F}$	& $\succsim^{2,T}$ & $\succsim^{2,F}$ & $\succsim^{2,F}$	 \\
			$S$4    & $\succsim^{2,T}$	  & $\succsim^{2,T}$	& $\succsim^{2,F}$	& $\succsim^{2,T}$ & $\succsim^{2,T}$\\
			$S$5    &$\succsim^{2,F}$($\succsim^{2,U}$)		  & $\succsim^{2,T}$	& $\succsim^{2,F}$($\succsim^{2,U}$)		& $\succsim^{2,F}$ & $\succsim^{2,T}$\\
			\hline
		\end{tabular}
	}
\end{table}


\begin{table}
	\caption{Outranking relations between students based on pairwise winning indices in the moderate perspective and outranking aggregation: $\succsim^{3,T}, \succsim^{3,F}$, and $\succsim^{3,U}$} \label{out_Moderate_out}
	\centering\footnotesize
	\resizebox{0.65\textwidth}{!}{
		\begin{tabular}{l|c|c|c|c|c}
			Student               & $S$1    & $S$2 & $S$3 & $S$4 & $S$5\\
			\hline
			$S$1    & $\succsim^{3,T}$	  & $\succsim^{3,T}$	& $\succsim^{3,T}$	& $\succsim^{3,F}$ & $\succsim^{3,T}$ \\
			$S$2    & $\succsim^{3,F}$	  & $\succsim^{3,T}$	& $\succsim^{3,T}$ & $\succsim^{3,F}$	& $\succsim^{3,T}$ \\
			$S$3    & $\succsim^{3,T}$	  & $\succsim^{3,F}$	& $\succsim^{3,T}$ & $\succsim^{3,F}$ & $\succsim^{3,F}$	 \\
			$S$4    & $\succsim^{3,T}$	  & $\succsim^{2,F}$($\succsim^{2,U}$)		& $\succsim^{3,F}$	& $\succsim^{3,T}$ & $\succsim^{2,F}$($\succsim^{2,U}$)	\\
			$S$5    & $\succsim^{3,T}$	  & $\succsim^{3,T}$	& $\succsim^{3,T}$	& $\succsim^{3,F}$ & $\succsim^{3,T}$\\
			\hline
		\end{tabular}
	}
\end{table}


Applying the ``corrected'' outranking relations $\succsim^{p,T},\ \succsim^{p,F}$, and $\succsim^{p,U}$, in all considered perspectives $p=1,2,3$, one can deduce in turn the overall seven-valued preference relations between students, presented in Table \ref{pref_seven-valued_corr}, where the original seven-valued outranking relations are put in parentheses when modified.

\begin{table}
	\caption{Overall seven-valued preference relations between students  ``corrected'' by pairwise winning indices in the three perspectives and outranking aggregation} \label{pref_seven-valued_corr}
	\centering\footnotesize
	\resizebox{0.65\textwidth}{!}{
		\begin{tabular}{l|c|c|c|c|c}
			Student               & $S$1    & $S$2 & $S$3 & $S$4 & $S$5\\
			\hline
			$S$1    & $\succsim^{T}$	  & $\succsim^{T}$	& $\succsim^{K}$($\succsim^{sT}$) 	& $\succsim^{F}$ & $\succsim^{T}$ \\
			$S$2    & $\succsim^{F}$	  & $\succsim^{T}$	& $\succsim^{K}$($\succsim^{sT}$) & $\succsim^{F}$	& $\succsim^{K}$($\succsim^{sT}$) \\
			$S$3    & $\succsim^{T}$	  & $\succsim^{F}$	& $\succsim^{T}$ & $\succsim^{F}$ & $\succsim^{F}$	 \\
			$S$4    & $\succsim^{T}$	  & $\succsim^{K}$($\succsim^{fK}$)	& $\succsim^{F}$	& $\succsim^{T}$ & $\succsim^{K}$($\succsim^{fK}$)\\
			$S$5    & $\succsim^{T}$	  & $\succsim^{T}$	& $\succsim^{K}$($\succsim^{sT}$)	& $\succsim^{F}$ & $\succsim^{T}$\\
			\hline
		\end{tabular}
	}
\end{table}

\vspace{6pt}

Using the ``basic'' values of gains and losses $v(S \succsim^{H} S')$, $v(S' \succsim^{H} S),\ H\in\{T,sT,U,K,fK,sF,F\}$, to the seven-valued outranking shown in Table \ref{pref_seven-valued_corr}, the five students were assigned the following global scores:
$$V^G(S1)=1, \ V^G(S2)=-3, \ V^G(S3)=-2, \ V^G(S4)=4, \ V^G(S5)=0,$$ 
resulting in the following ranking: $S4 \rightarrow S1 \rightarrow S5 \rightarrow S3 \rightarrow S2$.

%
\vspace{6pt}

Using the `deck-of-cards' method for finding values of gains and losses, in the same way as in the case of value function aggregation, the dean obtained the following global scores:
$$V^G(S1)=0.54, \ V^G(S2)=-2.08, \ V^G(S3)=-2, \ V^G(S4)=3.54, \ V^G(S5)=0,$$  
resulting in the same ranking of students as above.

%


\subsection{Incorporating indirect preference information via Robust Ordinal Regression and Stochastic Stochastic Multiobjective Acceptability Analysis}  

\vspace{6pt}
Suppose now that the dean would also like to express an indirect preference information in the form of holistic pairwise comparisons of some students in the three perspectives and see how the seven-valued preference relations and the final ranking would change. In particular, the dean provides the following pairwise comparisons: 
\begin{itemize}
	\item in the egalitarian perspective: 
	\begin{itemize}
		\item student $S2$ is at least as good as student $S3$ ($S2 \succsim_{DM}^1 S3$), and
		\item student $S4$ is at least as good as student $S3$ ($S4 \succsim_{DM}^1 S3$);
	\end{itemize}
	\item in the extreme perspective:t 
	\begin{itemize}
		\item student $S3$ is at least as good as student $S2$ ($S3 \succsim_{DM}^2 S2$), and
		\item student $S3$ is at least as good as student $S5$ ($S3 \succsim_{DM}^2 S5$);
	\end{itemize}
	\item in the moderate perspective: 
	\begin{itemize}
		\item student $S4$ is at least as good as student $S5$ ($S4 \succsim_{DM}^3 S5$), and
		\item student $S4$ is at least as good as student $S1$ ($S4 \succsim_{DM}^3 S1$).
	\end{itemize}
\end{itemize}

For each of the three perspectives, the set of weight vectors $\mathbf{\widetilde{w}}^p$ satisfying the preferences elicited from the dean must meet the following constraints:

\begin{equation}\label{ordinal regression}
	\begin{array}{l}
		\left.
		\begin{array}{l}
			\widetilde{w}^p_{Math} \geq 0,\ \widetilde{w}^p_{Phys} \geq 0,\ \widetilde{w}^p_{Lit}\geq 0,\ \widetilde{w}^p_{Phil}\geq 0,\\[1mm]
			\widetilde{w}^p_{Math}+\widetilde{w}^p_{Phys}+\widetilde{w}^p_{Lit}+\widetilde{w}^p_{Phil}=1,\\[1mm]
			U(S,\mathbf{\widetilde{w}}^p) \geqslant U(S',\mathbf{\widetilde{w}}^p) \mbox{ if }S\succsim_{DM}^p S',
		\end{array}
		\right\}E^p_{ (weight \; ordinal \; regression) }
	\end{array}\nonumber
\end{equation} 
where $S$ and $S'$ denote the students mentioned in the elicited preference information. The above constraints are typical for Robust Ordinal Regression introduced in \cite{ROR_originale,ROR}.
 
Our Propositions 1 and 2 also apply to the set of weight vectors compatible with preferences elicited from the dean and represented by constraints $E^p_{ (weight \; ordinal \; regression) }$. Thus, they can be used to compute the preference relations $\succsim^{p,T},\succsim^{p,F}$, and $\succsim^{p,U}$.

Based on Proposition 1, we present in Tables \ref{Ma_mi_egalitarian_or}, \ref{Ma_mi_extreme_or}, and \ref{Ma_mi_moderate_or}, the values of $m^p(S,S')$ and $M^p(S,S')$, and the resulting preference relation $\succsim^{p,H},\ H \in \{T,F,U\},\ p\in \{1,2,3\}$. As before, $m^p(S,S')$ and $M^p(S,S')$ denote the minimum and maximum values of compatible value functions  $U(S,\mathbf{\widetilde{w}}^p)-U(S',\mathbf{\widetilde{w}}^p)$, respectively, with $\mathbf{\widetilde{w}}^p \in E^p_{ (weight \; ordinal \; regression) }$. 

\begin{table}
	\caption{Values of $m^1(S,S')$ and $M^1(S,S')$, and the resulting preference relations between students in the egalitarian perspective for value functions obtained by ordinal regression: $\succsim^{1,T}$, $\succsim^{1,F}$, and $\succsim^{1,U}$} \label{Ma_mi_egalitarian_or}
	\centering\footnotesize
	\resizebox{1.05\textwidth}{!}{
		\begin{tabular}{l|c|c|c|c|c}
			Student               & $S$1    & $S$2 & $S$3 & $S$4 & $S$5\\
			\hline
			$S$1    & (0,0)$\rightarrow \succsim^{1,T}$	  & (-30,10)$\rightarrow \succsim^{1,U}$	& (-10,  30)$\rightarrow \succsim^{1,U}$	& (-10,7.5)$\rightarrow \succsim^{1,U}$ & (-20,10)$\rightarrow \succsim^{1,U}$ \\
			$S$2    & (-10,30)$\rightarrow\succsim^{1,U}$	  & (0,0)$\rightarrow\succsim^{1,T}$	& (0,30)$\rightarrow\succsim^{1,T}$ &(-14,20)$\rightarrow\succsim^{1,U}$	& (-4,10)$\rightarrow\succsim^{1,U}$ \\
			$S$3    & (-30,10)$\rightarrow\succsim^{1,U}$	  & (-30,0)$\rightarrow\succsim^{1,U}$	& (0,0)$\rightarrow\succsim^{1,T}$ & (-30,0)$\rightarrow\succsim^{1,U}$ & (-20,0)$\rightarrow\succsim^{1,U}$	 \\
			$S$4    & (-7.5,10)$\rightarrow\succsim^{1,U}$	  & (-20,14)$\rightarrow\succsim^{1,U}$	& (0,30)$\rightarrow\succsim^{1,T}$	& (0,0)$\rightarrow\succsim^{1,T}$ &(-10,10)$\rightarrow\succsim^{1,U}$\\
			$S$5    & (-10,20)$\rightarrow\succsim^{1,U}$	  & (-10,4)$\rightarrow\succsim^{1,U}$	& (0,20)$\rightarrow\succsim^{1,T}$	& (-10,10)$\rightarrow\succsim^{1,U}$ & (0,0)$\rightarrow\succsim^{1,T}$\\
			\hline
		\end{tabular}
	}
\end{table}


\begin{table}
	\caption{Values of $m^2(S,S')$ and $M^2(S,S')$, and the resulting preference relations between students in the extreme perspective for value functions obtained by ordinal regression: $\succsim^{2,T}$, $\succsim^{2,F}$, and $\succsim^{2,U}$} \label{Ma_mi_extreme_or}
	\centering\footnotesize
	\resizebox{1.05\textwidth}{!}{
		\begin{tabular}{l|c|c|c|c|c}
			Student               & $S$1    & $S$2 & $S$3 & $S$4 & $S$5\\
			\hline
			$S$1    & (0,0)$\rightarrow\succsim^{2,T}$	  & (-10,10)$\rightarrow\succsim^{2,U}$	& (-20,5)$\rightarrow\succsim^{2,U}$	& (-10,10)$\rightarrow\succsim^{2,U}$ & (-10,5)$\rightarrow\succsim^{2,U}$ \\
			$S$2    & (-10,-10)$\rightarrow\succsim^{2,U}$	  & (0,0)$\rightarrow\succsim^{2,T}$	& (-30,0)$\rightarrow\succsim^{2,U}$ & (-20,10)$\rightarrow\succsim^{2,U}$	& (-10,0)$\rightarrow\succsim^{2,U}$ \\
			$S$3    & (-5,20)$\rightarrow\succsim^{2,U}$	  & (0,30)$\rightarrow\succsim^{2,T}$	& (0,0)$\rightarrow\succsim^{2,T}$ & (-10,10)$\rightarrow\succsim^{2,U}$ & (0,20)$\rightarrow\succsim^{2,T}$	 \\
			$S$4    & (-10,10)$\rightarrow\succsim^{2,U}$	  & (-10,20)$\rightarrow\succsim^{2,U}$	& (-10,10)$\rightarrow\succsim^{2,U}$	& (0,0)$\rightarrow\succsim^{2,T}$ & (-10,10)$\rightarrow\succsim^{2,U}$\\
			$S$5    & (-5,10)$\rightarrow\succsim^{2,U}$	  & (0,10)$\rightarrow\succsim^{2,T}$	& (-20,0)$\rightarrow\succsim^{2,U}$	& (-10,10)$\rightarrow\succsim^{2,U}$ & (0,0)$\rightarrow\succsim^{2,T}$\\
			\hline
		\end{tabular}
	}
\end{table}


\begin{table}
	\caption{Values of $m^3(S,S')$ and $M^3(S,S')$, and the resulting preference relations between students in the moderate perspective for value functions obtained by ordinal regression: $\succsim^{3,T}, \succsim^{3,F}$, and $\succsim^{3,U}$} \label{Ma_mi_moderate_or}
	\centering\footnotesize
	\resizebox{1.05\textwidth}{!}{
		\begin{tabular}{l|c|c|c|c|c}
			Student               & $S$1    & $S$2 & $S$3 & $S$4 & $S$5\\
			\hline
			$S$1    & (0,0)$\rightarrow\succsim^{3,T}$	  & (-10,10)$\rightarrow\succsim^{3,U}$	& (-20,30)$\rightarrow\succsim^{3,U}$	&(-10,0)$\rightarrow\succsim^{3,U}$ &(-10,10)$\rightarrow\succsim^{3,U}$ \\
			$S$2    & (-10,10)$\rightarrow\succsim^{3,U}$	  &(0,0)$\rightarrow\succsim^{3,T}$	& (-30,25)$\rightarrow\succsim^{3,U}$ & (-20,5)$\rightarrow\succsim^{3,U}$	&(-10,5)$\rightarrow\succsim^{3,U}$ \\
			$S$3    &(-30,20)$\rightarrow\succsim^{3,U}$	  &(-25,30)$\rightarrow\succsim^{3,U}$	& (0,0)$\rightarrow\succsim^{3,T}$ &(-30,10)$\rightarrow\succsim^{3,U}$ &(-20,20)$\rightarrow\succsim^{3,U}$	 \\
			$S$4    &(0,10)$\rightarrow\succsim^{3,T}$	  &(-5,20)$\rightarrow\succsim^{3,U}$	&(-10,30)$\rightarrow\succsim^{3,U}$	&(0,0)$\rightarrow\succsim^{3,T}$ & (0,10)$\rightarrow\succsim^{3,T}$\\
			$S$5    & (-10,10)$\rightarrow\succsim^{3,U}$	  &(-5,10)$\rightarrow\succsim^{3,U}$	&(-20,20)$\rightarrow\succsim^{3,U}$	&(-10,0)$\rightarrow\succsim^{3,U}$ & (0,0)$\rightarrow\succsim^{3,T}$\\
			\hline
		\end{tabular}
	}
\end{table}


Based on Proposition 2, one can obtain three sets of vertex weight vectors compatible with the dean's preferences represented by constraints \linebreak $E^p_{ (weight \; ordinal \; regression)}$, $p\in\{1,2,3\}.$  These vertices are shown together with the corresponding overall evaluations of the five students in each of the considered perspectives in Tables \ref{Egalitarian_OR}, \ref{Extreme_OR}, and \ref{Moderate_OR}, respectively.

\begin{table}
	\caption{Vertex weight vectors and corresponding overall evaluations of students by value functions in the egalitarian perspective resulting from ordinal regression}\label{Egalitarian_OR}
	\centering\footnotesize
	\resizebox{1.0\textwidth}{!}{
		\begin{tabular}{l|cccc|ccccc}
			Weight  vector               & Mathematics    & Physics & Literature & Philosophy & S1 & S2 & S3 & S4 & S5 \\
			\hline
			$\mathbf{\widehat{w}}^{or,1,1}$ 	& 0	& 1	& 0	& 0 & 90 &       80&            60 &          90 &         80\\
			$\mathbf{\widehat{w}}^{or,1,2}$ 	& 0	& 0	& 1	& 0 & 50 &           80 &           50 &           60  &          70\\
			$\mathbf{\widehat{w}}^{or,1,3}$	& 0.4	& 0.6	& 0	& 0 & 86 &           76 &           76 &           90 &           80\\
			$\mathbf{\widehat{w}}^{or,1,4}$	& 0.5	& 0	& 0.5	& 0 & 65 &           75 &           75 &           75 &           75\\
			$\mathbf{\widehat{w}}^{or,1,5}$	& 0	& 0.25	& 0	& 0.75 & 75   &         72.5   &       67.5    &      67.5   &       72.5\\
			$\mathbf{\widehat{w}}^{or,1,6}$	& 0	& 0	& 0.5	& 0.5 & 60     &       75      &      60     &       60       &     70\\
			$\mathbf{\widehat{w}}^{or,1,7}$	& 0.17	& 0.25	& 0	& 0.58 & 76.67   &   72.5    &      72.5   &       72.5     &     74.17\\
			\hline
		\end{tabular}
	}
\end{table}

%
%
%
%
%
%
%

\begin{table}
	\caption{Vertex  weight vectors and corresponding overall evaluations of students by value functions in the extreme perspective resulting from ordinal regression}\label{Extreme_OR}
	\centering\footnotesize
	\resizebox{0.95\textwidth}{!}{
		\begin{tabular}{l|cccc|ccccc}
			Weight vector               & Mathematics    & Physics & Literature & Philosophy & S1 & S2 & S3 & S4 & S5 \\
			\hline
			$\mathbf{\widehat{w}}^{or, 2,    1            }$ &           1          &              0          &              0         &              0         &              80    &        70      &      100     &      90      &      80     \\            
			$\mathbf{\widehat{w}}^{or, 2,    2            }$ &           0          &              0          &              0         &              1         &              70          &             70         &             70      &             60      &             70          \\            
			$\mathbf{\widehat{w}}^{or, 2,    3            }$ &           0.5          &              0.5          &              0         &              0         &              85          &             75         &             80          &             90            &             80            \\            
			$\mathbf{\widehat{w}}^{or, 2,    4            }$ &           0.5          &              0          &              0.5         &              0         &              65          &             75      &             75          &             75            &             75            \\            
			\hline
		\end{tabular}
	}
\end{table}

%
%
%

\begin{table}
	\caption{Vertex  weight vectors and corresponding overall evaluations of students by value functions in the moderate perspective resulting from ordinal regression}\label{Moderate_OR}
	\centering\footnotesize
	\resizebox{0.95\textwidth}{!}{
		\begin{tabular}{l|cccc|ccccc}
			Weight vector               & Mathematics    & Physics & Literature & Philosophy & S1 & S2 & S3 & S4 & S5 \\
			\hline
			$\mathbf{\widehat{w}}^{or, 3,    1            }$ &           1        &              0         &              0          &              0          &              80         &    70   &          100    &        90     &        80      \\            
			$\mathbf{\widehat{w}}^{or, 3,    2            }$ &           0         &              1         &              0          &              0          &              90       &      80    &         60     &        90      &       80            \\            
			$\mathbf{\widehat{w}}^{or, 3,    3            }$ &           0.5         &              0         &              0.5          &              0          &              65        &     75        &     75     &        75     &        75     \\            
			$\mathbf{\widehat{w}}^{or, 3,    4            }$ &           0.5         &              0         &              0          &              0.5          &              75       &      70       &      85       &      75     &        75      \\            
			$\mathbf{\widehat{w}}^{or, 3,    5            }$ &           0         &              0.5         &              0.5          &              0          &              70       &      80      &       55     &        75   &          75           \\            
			$\mathbf{\widehat{w}}^{or, 3,    6            }$ &           0         &              0.5         &              0.25          &              0.25          & 75      &       77.5      &     60       &      75      &       75      \\            
			\hline
		\end{tabular}
	}
\end{table}
%
%
%
%
%
%
%
%
Taking into account the preference relations $\succsim^{p,T},\ \succsim^{p,F}$, and $\succsim^{p,U}$, in all considered perspectives $p=1,2,3$, presented in Tables \ref{Ma_mi_egalitarian_or}, \ref{Ma_mi_extreme_or}, and \ref{Ma_mi_moderate_or}, one can deduce in turn the overall seven-valued preference relations between students, presented in Table \ref{pref_seven_valued_or_vf}. 

\begin{table}
	\caption{Overall seven-valued preference relations between students resulting from value function aggregation and ordinal regression}\label{pref_seven_valued_or_vf}
	\centering\footnotesize
	\resizebox{0.45\textwidth}{!}{
		\begin{tabular}{l|ccccc}
			Student               & $S$1    & $S$2 & $S$3 & $S$4 & $S$5\\
			\hline
			$S$1    & $\succsim^{T}$	  & $\succsim^{U}$	& $\succsim^{U}$	& $\succsim^{U}$ & $\succsim^{U}$ \\
			$S$2    & $\succsim^{U}$	  & $\succsim^{T}$	& $\succsim^{sT}$ & $\succsim^{U}$	& $\succsim^{U}$ \\
			$S$3    & $\succsim^{U}$	  & $\succsim^{sT}$	& $\succsim^{T}$ & $\succsim^{U}$ & $\succsim^{sT}$	 \\
			$S$4    & $\succsim^{sT}$	  & $\succsim^{U}$	& $\succsim^{sT}$	& $\succsim^{T}$ & $\succsim^{sT}$\\
			$S$5    & $\succsim^{U}$	  & $\succsim^{sT}$	& $\succsim^{sT}$	& $\succsim^{U}$ & $\succsim^{T}$\\
			\hline
		\end{tabular}
	}
\end{table}

\vspace{6pt}

Applying the ``basic'' values of the gains and losses \linebreak $v(S \succsim^{H} S'),\ v(S' \succsim^{H} S),\ H\in\{T,sT,U,K,fK,sF,F\}$, to the seven-valued preference relations shown in Table \ref{pref_seven_valued_or_vf}, the five students were assigned the following global scores:
$$V^G(S1)=-0.5, \ V^G(S2)=-0.5, \ V^G(S3)=-0.5, \ V^G(S4)=1.5, \ V^G(S5)=0,$$ 
resulting in the following ranking: $S4 \rightarrow S5 \rightarrow S1 \sim S2 \sim S3$.


\vspace{6pt}

Using the `deck-of-cards' method for finding values of gains and losses, the dean obtained the following global scores:
$$V^G(S1)=0.54, \ V^G(S2)=-2.08, \ V^G(S3)=-2, \ V^G(S4)=3.54, \ V^G(S5)=0,$$  
resulting in the same ranking of students as above.


\vspace{6pt}
To avoid bias in the seven-valued preference relations resulting from overall evaluations by value functions with weight vectors located only at the vertices of $E^p_{(weight \; ordinal \; regression)}$, the dean considered the probability $Pr(S \succsim S')$ of student $S$ being preferred over student $S'$. These probabilities, called ``pairwise winning indices'', were obtained with a methodology called Stochastic Ordinal Regression \cite{SOR}, as above, using SMAA with a uniform probability distribution in the space of feasible weights, and, more precisely, using the `hit-and-run' algorithm in the simplex $E^p_{(weight \; ordinal \; regression)}$ with a random sampling of 100,000 weight vectors for each perspective $p=1,2,3$. The results obtained for the three perspectives are shown in Tables \ref{SMAA_Egalitarian_or}, \ref{SMAA_Extreme_or}, \ref{SMAA_Moderate_or}, respectively.

\begin{table}
	\caption{Pairwise winning indices of students in rows over students in columns in the egalitarian perspective and value functions obtained by ordinal regression and SMAA} \label{SMAA_Egalitarian_or}
	\centering\footnotesize
	\resizebox{0.45\textwidth}{!}{
		\begin{tabular}{l|ccccc}
			Student               & $S$1    & $S$2 & $S$3 & $S$4 & $S$\\
			\hline
			$S$1    & 1	  & 0.37	& 0.79	& 0.23 & 0.38 \\
			$S$2    & 0.63	  & 1	& 1 & 0.57	& 0.66 \\
			$S$3    & 0.21	  & 0	& 1 & 0 & 0	 \\
			$S$4    & 0.77	  & 0.43	& 1	& 1 & 0.47\\
			$S$5    & 0.62	  & 0.33	& 1	& 0.53 & 1\\
			\hline
		\end{tabular}
	}
\end{table}


\begin{table}
	\caption{Pairwise winning indices of students in rows over students in columns in the extreme perspective and value functions obtained by ordinal regression and SMAA} \label{SMAA_Extreme_or}
	\centering\footnotesize
	\resizebox{0.45\textwidth}{!}{
		\begin{tabular}{l|ccccc}
			Student               & $S$1    & $S$2 & $S$3 & $S$4 & $S$\\
			\hline
			$S$1    & 1	  & 0.72	& 0.11	& 0.26 & 0.39 \\
			$S$2    & 0.27	  & 1	& 0 & 0.19	& 0 \\
			$S$3    & 0.89	  & 1	& 1 & 0.74 & 1	 \\
			$S$4    & 0.77	  & 0.81	& 0.251	& 1 & 0.70\\
			$S$5    & 0.61	  & 1	& 0	& 0.30 & 1\\
			\hline
		\end{tabular}
	}
\end{table}

\begin{table}
	\caption{Pairwise winning indices of students in rows over students in columns in the moderate perspective and value functions obtained by ordinal regression and SMAA} \label{SMAA_Moderate_or}
	\centering\footnotesize
	\resizebox{0.45\textwidth}{!}{
		\begin{tabular}{l|ccccc}
			Student               & $S$1    & $S$2 & $S$3 & $S$4 & $S$\\
			\hline
			$S$1    & 1	  & 0.66	& 0.54	& 0 & 0.5 \\
			$S$2    & 0.34	  & 1	& 0.51 & 0.12	& 0.27 \\
			$S$3    & 0.46	  & 0.49	& 1 & 0.31 & 0.43	 \\
			$S$4    & 1	  & 0.88	& 0.69	& 1 & 1\\
			$S$5    & 0.5	  & 0.73	& 0.57	& 0 & 1\\
			\hline
		\end{tabular}
	}
\end{table}

Taking into account the pairwise winning indices from Tables \ref{SMAA_Egalitarian_or}, \ref{SMAA_Extreme_or}, and \ref{SMAA_Moderate_or}, and setting again a threshold of  $t=0.85$ on these probabilities,  the true, false, and unknown preference relations, $\succsim^{p,T},\ \succsim^{p,F}$, and $\succsim^{p,U},\ p\in\{1,2,3\}$, are shown in Tables \ref{pref_Egalitarian_or_}, \ref{pref_Extreme_or_} and \ref{pref_Moderate_or_}, where the original values are put in parentheses when modified.

\begin{table}
	\caption{Preference relations between students based on pairwise winning indices in the egalitarian perspective and value functions obtained by ordinal regression and SMAA: $\succsim^{1,T}, \succsim^{1,F}$, and $\succsim^{1,U}$} \label{pref_Egalitarian_or_}
	\centering\footnotesize
	\resizebox{0.75\textwidth}{!}{
		\begin{tabular}{l|c|c|c|c|c}
			Student               & $S$1    & $S$2 & $S$3 & $S$4 & $S$5\\
			\hline
			$S$1    & $\succsim^{1,T}$	  & $\succsim^{1,U}$	& $\succsim^{1,U}$	& $\succsim^{1,U}$ & $\succsim^{1,U}$ \\
			$S$2    & $\succsim^{1,U}$	  & $\succsim^{1,T}$	& $\succsim^{1,T}$ & $\succsim^{1,U}$	& $\succsim^{1,U}$ \\
			$S$3    & $\succsim^{1,U}$	  & $\succsim^{1,F}$($\succsim^{1,U}$)& $\succsim^{1,T}$ & $\succsim^{1,F}$($\succsim^{1,U}$)& $\succsim^{1,F}$($\succsim^{1,U}$)\\
			$S$4    & $\succsim^{1,U}$	  & $\succsim^{1,U}$	& $\succsim^{1,T}$	& $\succsim^{1,T}$ & $\succsim^{1,U}$\\
			$S$5    & $\succsim^{1,U}$	  & $\succsim^{1,U}$	& $\succsim^{1,T}$	& $\succsim^{1,U}$ & $\succsim^{1,T}$\\
			\hline
		\end{tabular}
	}
\end{table}


\begin{table}
	\caption{Preference relations between students based on pairwise winning indices in the extreme perspective and value functions obtained by ordinal regression and SMAA: $\succsim^{2,T}, \succsim^{2,F}$, and $\succsim^{2,U}$} \label{pref_Extreme_or_}
	\centering\footnotesize
	\resizebox{0.75\textwidth}{!}{
		\begin{tabular}{l|c|c|c|c|c}
			Student               & $S$1    & $S$2 & $S$3 & $S$4 & $S$5\\
			\hline
			$S$1    & $\succsim^{2,T}$	  & $\succsim^{2,U}$	& $\succsim^{2,F}$($\succsim^{2,U}$) & $\succsim^{2,U}$ & $\succsim^{2,U}$ \\
			$S$2    & $\succsim^{2,U}$	  & $\succsim^{2,T}$	& $\succsim^{2,F}$($\succsim^{2,U}$)& $\succsim^{2,U}$	& $\succsim^{2,F}$($\succsim^{3,U}$)\\
			$S$3    & $\succsim^{2,T}$($\succsim^{2,U}$) & $\succsim^{2,T}$ 	& $\succsim^{2,T}$ & $\succsim^{2,U}$ & $\succsim^{2,T}$	 \\
			$S$4    & $\succsim^{2,U}$	  & $\succsim^{2,U}$	& $\succsim^{2,U}$	& $\succsim^{2,T}$ & $\succsim^{2,U}$\\
			$S$5    & $\succsim^{2,U}$	 & $\succsim^{2,T}$	& $\succsim^{2,F}$($\succsim^{2,U}$) & $\succsim^{2,U}$ & $\succsim^{2,T}$\\
			\hline
		\end{tabular}
	}
\end{table}


\begin{table}
	\caption{Preference relations between students based on pairwise winning indices in the moderate perspective and value functions obtained by ordinal regression and SMAA: $\succsim^{3,T}, \succsim^{3,F}$, and $\succsim^{3,U}$} \label{pref_Moderate_or_}
	\centering\footnotesize
	\resizebox{0.7\textwidth}{!}{
		\begin{tabular}{l|c|c|c|c|c}
			Student               & $S$1    & $S$2 & $S$3 & $S$4 & $S$5\\
			\hline
			$S$1    & $\succsim^{3,T}$	  & $\succsim^{3,U}$	& $\succsim^{3,U}$	& $\succsim^{3,F}$($\succsim^{3,U}$)& $\succsim^{3,U}$  \\
			$S$2    & $\succsim^{3,U}$	  & $\succsim^{3,T}$	& $\succsim^{3,U}$  & $\succsim^{3,F}$($\succsim^{3,U}$)& $\succsim^{3,U}$ \\
			$S$3    & $\succsim^{3,U}$	  & $\succsim^{3,U}$ 	& $\succsim^{3,T}$ & $\succsim^{3,U}$ & $\succsim^{3,U}$	 \\
			$S$4    & $\succsim^{3,T}$	  & $\succsim^{3,T}$($\succsim^{3,U}$)& $\succsim^{3,U}$	& $\succsim^{3,T}$ & $\succsim^{3,T}$\\
			$S$5    & $\succsim^{3,U}$	  & $\succsim^{3,U}$	& $\succsim^{3,U}$	& $\succsim^{3,F}$($\succsim^{3,U}$)& $\succsim^{3,T}$\\
			\hline
		\end{tabular}
	}
\end{table}


Applying the ``corrected'' outranking relations $\succsim^{p,T},\ \succsim^{p,F}$, and $\succsim^{p,U}$, in all considered perspectives $p=1,2,3$,  presented in Tables \ref{pref_Egalitarian_or_}, \ref{pref_Extreme_or_}, and \ref{pref_Moderate_or_}, one can deduce in turn the overall seven-valued preference relations between students, presented in Table \ref{pref_seven-valued_or_}, where the original seven-valued preference relations are put in parentheses when modified. 

%
\vspace{6pt}



\begin{table}
	\caption{Overall seven-valued preference relations between students resulting from value function aggregation, ordinal regression and SMAA} \label{pref_seven-valued_or_}
	\centering\footnotesize
	\resizebox{0.8\textwidth}{!}{
		\begin{tabular}{l|c|c|c|c|c}
			Student               & $S$1    & $S$2 & $S$3 & $S$4 & $S$5\\
			\hline
			$S$1    & $\succsim^{T}$	  & $\succsim^{U}$	& $\succsim^{sF}$	($\succsim^{U}$)& $\succsim^{sF}$ ($\succsim^{U}$)& $\succsim^{U}$ \\
			$S$2    & $\succsim^{U}$	  & $\succsim^{T}$	& $\succsim^{fK}$ ($\succsim^{sT}$)& $\succsim^{sF}$	($\succsim^{U}$)& $\succsim^{sF}$ ($\succsim^{U}$)\\
			$S$3    & $\succsim^{sT}$	  & $\succsim^{fK}$	($\succsim^{sT}$)& $\succsim^{T}$ & $\succsim^{sF}$ ($\succsim^{U}$)& $\succsim^{fK}$	($\succsim^{sT}$) \\
			$S$4    & $\succsim^{sT}$	  & $\succsim^{sT}$	($\succsim^{U}$)& $\succsim^{sT}$	& $\succsim^{T}$ & $\succsim^{sT}$\\
			$S$5    & $\succsim^{U}$	  & $\succsim^{sT}$	& $\succsim^{fK}$	($\succsim^{sT}$)& $\succsim^{sF}$ ($\succsim^{U}$)& $\succsim^{T}$\\
			\hline
		\end{tabular}
	}
\end{table}

\vspace{6pt}
Applying the ``basic'' values of the gains and losses \linebreak $v(S \succsim^{H} S'),\ v(S' \succsim^{H} S),\ H\in\{T,sT,U,K,fK,sF,F\}$, to the seven-valued preference relations shown in Table \ref{pref_seven-valued_or_}, the five students were assigned the following global scores:
$$V^G(S1)=-2, \ V^G(S2)=-2, \ V^G(S3)=0, \ V^G(S4)=4, \ V^G(S5)=0,$$ 
resulting in the following ranking:  $S4 \rightarrow S3 \sim S5 \rightarrow  S1 \sim S2$.

Using the `deck-of-cards' method for finding values of gains and losses, the dean obtained the following global scores:
$$V^G(S1)=-0.31, \ V^G(S2)=-0.31, \ V^G(S3)=-0.31, \ V^G(S4)=0.92, \ V^G(S5)=0,$$  
resulting in the following ranking of students: $S4 \rightarrow S5 \rightarrow  S1 \sim  S2 \sim S3$.

\section{Conclusions}

Each multiple criteria decision aiding procedure requires constructing a decision model that respects the preferences of the decision maker. This can only be achieved through collaboration between the analyst and the decision maker. Assigning values to the preference parameters of the decision model is crucial for the credibility of the final recommendation. These parameters do not have objectively true values, so it is reasonable to explore the feasible space of preference parameters from several perspectives and consider reasonable perturbations around their central values.

This exploration allows one to express preference relations among alternatives using a seven-valued logic, which we introduced in this paper to enhance its natural and straightforward derivation. We demonstrated that the seven-valued preference structure can be applied throughout the decision aiding procedure. This includes defining different perspectives for adopting preference parameter values, constructing and explaining the seven-value preferences, and using these preferences to make appropriate recommendations.

Our proposed methodology can be applied to both value function aggregation and outranking aggregation. It incorporates and systematizes recent developments in MCDA, including stochastic multiobjective acceptability analysis, robust ordinal regression, and robust ordinal regression with stochastic multiobjective acceptability analysis.

For future research, we plan to explore the use of specific forms of value functions such as the Choquet integral \cite{Choquet}, or outranking functions used in PROMETHEE methods \cite{Promethee}. Additionally, we aim to apply this methodology to robust multiobjective optimization.

\subsubsection*{Acknowledgments.} Salvatore Greco wishes to acknowledge the support of the Ministero dell’Istruzione, dell’Universit\'{a} e della Ricerca (MIUR) - PRIN 2017, project ``Multiple Criteria Decision Analysis and Multiple Criteria Decision Theory'', grant 2017CY2NCA. The research of Roman  S\l owi\'nski
was supported by the SBAD funding from the Polish Ministry of Education and Science. This research also contributes to the PNRR GRInS Project.

\appendix
\section*{Appendix A}
 
\textbf{Proof of Proposition 1.} For all pairs of students, $S$ and $S'$, $S \succsim^{p,T} S'$ if and only if $U(S,\mathbf{\widetilde{w}}^p) \geqslant U(S',\mathbf{\widetilde{w}}^p)$ for all $\mathbf{\widetilde{w}}^p \in E^p_{ (wp)}$, which is equivalent to $m^p(S,S') \geqslant 0$,  where $m^p(S,S')=min [U(S)-U(S')]$ subject to $E^p_{ (wp)}$. Analogously, $S \succsim^{p,F} S'$ if and only if $U(S,\mathbf{\widetilde{w}}^p) < U(S',\mathbf{\widetilde{w}}^p)$ for all $\mathbf{\widetilde{w}}^p \in E^p_{ (wp)}$, which is equivalent to $M^p(S,S') < 0$,  where $M^p(S,S')=max [U(S)-U(S')]$ subject to $E^p_{ (wp)}$. Finally, $S \succsim^{p,U} S'$ is equivalent to existence of a weight vector $\mathbf{\widetilde{w}}^{p,1} \in E^p_{ (wp)}$ for which $U(S) \geqslant U(S')$, as well as existence of another weight vector $\mathbf{\widetilde{w}}^{p,2} \in E^p_{ (wp)}$ for which $U(S) < U(S')$. Taking $\mathbf{\widetilde{w}}^{p,1}$ and $\mathbf{\widetilde{w}}^{p,2}$ as the weight vectors for which $U(S,\mathbf{\widetilde{w}}^{p,1})-U(S',\mathbf{\widetilde{w}}^{p,1})=M^p(S,S')$ and $U(S,\mathbf{\widetilde{w}}^{p,2})-U(S',\mathbf{\widetilde{w}}^{p,2})=m^p(S,S')$, we have that $S \succsim^{p,U} S'$ is equivalent to $m^p(S,S') < 0 \leqslant M^p(S,S')$. $\square$

\appendix
\section*{Appendix B}

\textbf{Proof of Proposition 2.} Let us prove that $S \succsim^{p,T} S'$ implies $U(S,\mathbf{\widetilde{w}}^p) \geq U(S',\mathbf{\widetilde{w}}^p)$ for all $\mathbf{\widetilde{w}}^p \in V(E^p_{ (wp)})$. Suppose that $S \succsim^{p,T} S'$. In this case, by definition,  $U(S,\mathbf{\widetilde{w}}^p) \geq U(S',\mathbf{\widetilde{w}}^p)$ for all $\mathbf{\widetilde{w}}^p \in E^p_{ (w,p) }$, which implies that $U(S,\mathbf{\widetilde{w}}^p) \geq U(S',\mathbf{\widetilde{w}}^p)$ for all $\mathbf{\widetilde{w}}^p \in V(E^p_{ (wp)})$ because, clearly,  $V(E^p_{ (wp)}) \subseteq E^p_{ (w,p) }$. 

Let us prove, in turn, that $U(S,\mathbf{\widetilde{w}}^p) \geq U(S',\mathbf{\widetilde{w}}^p)$ for all $\mathbf{\widetilde{w}}^p \in V(E^p_{ (wp)})$ implies $S \succsim^{p,T} S'$. Suppose that $U(S,\mathbf{\widetilde{w}}^p) \geq U(S',\mathbf{\widetilde{w}}^p)$ for all $\mathbf{\widetilde{w}}^p \in V(E^p_{ (wp)})$. Since for all $\mathbf{\widetilde{w}}^p \in E^p_{ (wp)}$ there exists a vector $\alpha_{{\mathbf{\widehat{w}}}}=[\alpha_{\mathbf{\widehat{w}}}^p, \widehat{w}^p \in V(E^p_{ (wp)}]$ with $\alpha_{{\mathbf{\widehat{w}}}^p} \geq 0$ for all vertices $\mathbf{\widehat{w}}^p \in V(E^p_{(wp)})$ and $\sum_{{\mathbf{\widehat{w}}^p}\in V(E^p_{ (wp)})} \alpha_{\mathbf{\widehat{w}}^p}=1$, such that 
\[
\mathbf{\widetilde{w}}^p=\sum_{{\mathbf{\widehat{w}}^p}\in V(E^p_{ (wp)})} \alpha_{\mathbf{\widehat{w}}^p}\times \mathbf{\widehat{w}}^p
\]
for all student $\overline{S}$, we have
\[
U(\overline{S},\mathbf{\widetilde{w}}^p)=\sum_{s_j\in \mathcal{S}}\widetilde{w}^p_{s_j}g_{s_j}(\overline{S})=\sum_{s_j\in \mathcal{S}}\left(\sum_{{\mathbf{\widehat{w}}^p}\in V(E^p_{ (wp)})} \alpha_{\mathbf{\widehat{w}}^p}\times \mathbf{\widehat{w}}^p_{s_j}\right)g_{s_j}(\overline{S})=
\]
\begin{equation}\label{Value_vertices}
	\sum_{{\mathbf{\widehat{w}}^p}\in V(E^p_{ (wp)})} \alpha_{\mathbf{\widehat{w}}^p}\times\left(\sum_{s_j\in \mathcal{\overline{S}}} \mathbf{\widehat{w}}^p_{s_j} \times g_{s_j}(\overline{S})\right)=\sum_{{\mathbf{\widehat{w}}^p}\in V(E^p_{ (wp)})} \alpha_{\mathbf{\widehat{w}}^p}\times U(\overline{S},\mathbf{\widehat{w}}^p)
\end{equation}
with $\mathcal{S}=\{Math,Phys,Lit,Phil\}$.
Taking into account equation (\ref{Value_vertices}), from 
$$U(S,\mathbf{\widetilde{w}}^p) \geq U(S',\mathbf{\widetilde{w}}^p)$$
we get  that for all $\mathbf{\widehat{w}}^p \in E^p_{ (wp)}$,
\[
U(S,\mathbf{\widehat{w}}^p)=\sum_{{\mathbf{\widehat{w}}^p}\in V(E^p_{ (wp)})} \alpha_{\mathbf{\widehat{w}}^p}\times U(S,\mathbf{\widehat{w}}^p) \geqslant \sum_{{\mathbf{\widehat{w}}^p}\in V(E^p_{ (wp)})} \alpha_{\mathbf{\widehat{w}}^p}\times U(S',\mathbf{\widehat{w}}^p)=U(S',\mathbf{\widehat{w}}^p),
\]
which implies, by definition, that $S \succsim^{p,T} S'$.

Thus we proved that $S \succsim^{p,T} S'$ if and only if  $U(S,\mathbf{\widetilde{w}}^p) \geq U(S',\mathbf{\widetilde{w}}^p)$ for all $\mathbf{\widetilde{w}}^p \in V(E^p_{ (wp)})$. Analogously, one can prove that $S \succsim^{p,F} S'$ if and only if  $U(S,\mathbf{\widetilde{w}}^p) < U(S',\mathbf{\widetilde{w}}^p)$ for all $\mathbf{\widetilde{w}}^p \in V(E^p_{ (wp)})$.

Now, let us prove that $S \succsim^{p,U} S'$ implies $U(S,\mathbf{\widetilde{w}}^p) \geq U(S',\mathbf{\widetilde{w}}^p)$ for some $\mathbf{\widetilde{w}}^p \in V(E^p_{ (wp)})$ and $U(S,\mathbf{\widetilde{w}}^p) < U(S',\mathbf{\widetilde{w}}^p)$ for some other $\mathbf{\widetilde{w}}^p \in V(E^p_{ (wp)})$. By contradiction, suppose that 
$S \succsim^{p,U} S'$ and $U(S,\mathbf{\widetilde{w}}^p) < U(S',\mathbf{\widetilde{w}}^p)$ for all $\mathbf{\widetilde{w}}^p \in V(E^p_{ (wp)})$.
Taking into account equation (\ref{Value_vertices}), from $U(S,\mathbf{\widetilde{w}}^p) < U(S',\mathbf{\widetilde{w}}^p)$ for all $\mathbf{\widetilde{w}}^p \in V(E^p_{ (wp)})$,  we would get
\[
U(S,\mathbf{\widehat{w}}^p)=\sum_{{\mathbf{\widehat{w}}^p}\in V(E^p_{ (wp)})} \alpha_{\mathbf{\widehat{w}}^p}\times U(S,\mathbf{\widehat{w}}^p) < \sum_{{\mathbf{\widehat{w}}^p}\in V(E^p_{ (wp)})} \alpha_{\mathbf{\widehat{w}}^p}\times U(S',\mathbf{\widehat{w}}^p)=U(S',\mathbf{\widehat{w}}^p)
\]
for all $\mathbf{\widehat{w}}^p \in E^p_{ (wp)}$, which should lead to conclusion $S \succsim^{p,F} S'$, rather than  $S \succsim^{p,U} S'$, which is absurd. 
Analogously, again by contradiction, supposing that 
$S \succsim^{p,U} S'$ and $U(S,\mathbf{\widetilde{w}}^p) \geqslant U(S',\mathbf{\widetilde{w}}^p)$ for all $\mathbf{\widetilde{w}}^p \in V(E^p_{ (wp)})$, one would get 
\[
U(S,\mathbf{\widehat{w}}^p) \geqslant U(S',\mathbf{\widehat{w}}^p)
\]
for all $\mathbf{\widehat{w}}^p \in E^p_{ (wp)}$, which should lead to conclusion $S \succsim^{p,T} S'$, rather than  $S \succsim^{p,U} S'$, which is absurd. Consequently, we have to conclude that if $S \succsim^{p,U} S'$, then $U(S,\mathbf{\widetilde{w}}^p) \geq U(S',\mathbf{\widetilde{w}}^p)$ for some $\mathbf{\widetilde{w}}^p \in V(E^p_{ (wp)})$ and $U(S,\mathbf{\widetilde{w}}^p) < U(S',\mathbf{\widetilde{w}}^p)$ for some other $\mathbf{\widetilde{w}}^p \in V(E^p_{ (wp)})$.

Note that if $U(S,\mathbf{\widetilde{w}}^p) \geq U(S',\mathbf{\widetilde{w}}^p)$  for some $\mathbf{\widetilde{w}}^p \in V(E^p_{ (wp)})$ and $U(S,\mathbf{\widetilde{w}}^p) < U(S',\mathbf{\widetilde{w}}^p)$ for some other $\mathbf{\widetilde{w}}^p \in V(E^p_{ (wp)})$, by definition, $S \succsim^{p,U} S'$ because, clearly, $V(E^p_{ (wp)}) \subseteq E^p_{ (w,p) }$. $\square$

\end{document}